\documentclass[letterpaper]{article} 
\usepackage{aaai2026}  
\usepackage{times}  
\usepackage{helvet}  
\usepackage{courier}  
\usepackage[hyphens]{url}  
\usepackage{graphicx} 
\usepackage{natbib}  
\usepackage{caption} 
\usepackage{algorithm}
\usepackage{booktabs}
\usepackage{algorithmic}
\usepackage{amsmath}
\usepackage{amssymb}
\usepackage{mathtools}
\usepackage{amsthm}
\usepackage{multirow}

\newtheorem{definition}{Definition}

\usepackage{newfloat}
\usepackage{listings}
\DeclareCaptionStyle{ruled}{labelfont=normalfont,labelsep=colon,strut=off} 
\floatstyle{ruled}
\newfloat{listing}{tb}{lst}{}
\floatname{listing}{Listing}

\usepackage{listings}

\lstdefinestyle{pythonstyle}{
  language=Python,
  basicstyle=\ttfamily\scriptsize,  
  numbers=none,
  numberstyle=\tiny\color{gray},
  stepnumber=1,
  numbersep=5pt,
  frame=single,
  rulecolor=\color{black},
  tabsize=2,
  captionpos=b,
  breaklines=true,
  breakatwhitespace=false,  
  keywordstyle=\color{blue},
  commentstyle=\color{teal},
  stringstyle=\color{red},
  columns=flexible,  
  xleftmargin=\fill,   
  xrightmargin=\fill   
}

\lstdefinestyle{pythonstylesmall}{
  language=Python,
  basicstyle=\ttfamily\scriptsize,  
  numbers=none,
  numberstyle=\tiny\color{gray},
  stepnumber=1,
  numbersep=5pt,
  frame=single,
  rulecolor=\color{black},
  tabsize=2,
  captionpos=b,
  breaklines=true,
  breakatwhitespace=false,  
  keywordstyle=\color{blue},
  commentstyle=\color{teal},
  stringstyle=\color{red},
  columns=flexible,  
  xleftmargin=\fill,   
  xrightmargin=\fill   
}

\lstdefinestyle{text}{
  basicstyle=\ttfamily\scriptsize,
  breaklines=true,
  breakindent=0pt,
  breakautoindent=false,
  breakatwhitespace=true,
  columns=fullflexible,
  frame=single,
  xleftmargin=0pt,
  xrightmargin=0pt
}

\lstdefinestyle{textsmall}{
  basicstyle=\ttfamily\scriptsize,
  breaklines=true,
  breakindent=0pt,
  breakautoindent=false,
  breakatwhitespace=true,
  columns=fullflexible,
  frame=single,
  xleftmargin=0pt,
  xrightmargin=0pt,
  aboveskip=2pt,
  belowskip=2pt,
  lineskip=-1pt
}

\lstdefinelanguage{PDDL}{
    comment=[l]{;},
    commentstyle=\color{gray}\itshape,
    morestring=[b]",
    stringstyle=\color{red},
    sensitive=true
}

\lstdefinestyle{pddlstyle}{
    language=PDDL,
    basicstyle=\ttfamily\scriptsize, 
    columns=fullflexible,
    keepspaces=true,
    numbers=none, 
    showstringspaces=false,
    tabsize=2,
    breaklines=true,
    breakatwhitespace=false,
    frame=none, 
}
\title{GenePlan: Evolving Better Generalized PDDL Plans using Large Language Models}
\author{
    Andrew Murray,
    Danial Dervovic,
    Alberto Pozanco,
    Michael Cashmore
}
\affiliations{
    J.P. Morgan AI Research\\

    \{andrew.murray, danial.dervovic, alberto.pozancolancho, michael.cashmore\}@jpmorgan.com
}

\usepackage{bibentry}

\begin{document}

\maketitle

\begin{abstract}
We present GenePlan (GENeralized Evolutionary Planner), a novel framework that leverages large language model (LLM) assisted evolutionary algorithms to generate domain-dependent generalized planners for classical planning tasks described in PDDL. By casting generalized planning as an optimization problem, GenePlan iteratively evolves interpretable Python planners that minimize plan length across diverse problem instances. In empirical evaluation across six existing benchmark domains and two new domains, GenePlan achieved an average SAT score of 0.91, closely matching the performance of the state-of-the-art planners (SAT score 0.93), and significantly outperforming other LLM-based baselines such as chain-of-thought (CoT) prompting (average SAT score 0.64). The generated planners solve new instances rapidly (average 0.49 seconds per task) and at low cost (average \$1.82 per domain using GPT-4o).
\end{abstract}

\section{Introduction}
Large language models (LLMs) are neural networks that have been trained on vast amounts of data and have been applied successfully to a broad spectrum of tasks: from question-answering to writing code \cite{bubeck2023sparks}. Despite recent advances in LLM reasoning \cite{guo2025deepseek}, their use in sequential decision making tasks such as planning has thus far shown sub-par performance \cite{valmeekam2022large}.

Recent advances in the field of \textit{generalized planning} \cite{jimenez2019review} has shown promise, whereby purpose built Python planners are designed by the LLM to solve new instances in a given domain \cite{silver2024generalized}. Generating such solutions requires significant domain knowledge, a task that the LLM is well suited to, given its vast training set. However these approaches focus on generating satisficing solutions with no consideration for plan quality. This is insufficient in many practical applications where the quality of the solution matters.

In tandem, there has been some success in using LLMs as \textit{optimizers}, where the optimization problem and its solution are defined in natural language \cite{yang2023large}. \citet{romera2024mathematical} build upon this by embedding an LLM within an evolutionary optimization framework. In this approach, the solution is a Python method and the optimization space is any valid Python code. The authors applied this framework to generate new heuristic functions for combinatorial optimization problems and found that the LLM was able to come up with heuristics surpassing the best human-generated heuristic.

In this paper, we build upon both these recent lines of work. In particular, we extend the chain-of-thought (CoT) approach taken by \citet{silver2024generalized}, to use the evolutionary LLM framework introduced by~\citet{romera2024mathematical}.

Our method, GenePlan, treats generalized planning as an \textit{optimization} problem. The goal is to produce a generalized plan, written in Python, that minimizes the number of actions across a training set of planning tasks. GenePlan iteratively develops a population of candidate Python methods (generalized plans). The LLM proposes new methods, which are evaluated for plan quality. These methods are stored in the population and fed back to the LLM through evolutionary sampling. The population evolves, with the lowest-performing members pruned each generation, converging on high-quality solutions. The optimal solution, an efficient Python method, is extracted at the end, capable of generating \textit{high-quality} and \textit{interpretable} solutions for new PDDL planning tasks in the domain.

We empirically evaluate our approach on 8 domains, comparing 10 baseline approaches and show that our approach is comparable in terms of plan quality with state of the art planners given a 30 minute time limit). The resulting Python planner is interpretable and is capable of solving new planning instances rapidly (an average of 0.49 seconds per task). The average cost for generating a planner using our approach was just \$1.82 per domain.

In section \ref{sec:bg} we describe the background relevant to this work.
In section \ref{sec:rw} we place the contribution of this paper in context with related work.
In section~\ref{sec:method} we outline the GenePlan procedure.
In section~\ref{sec:experiments} and \ref{sec:results} we describe the setup and results of our experimental evaluation.
We conclude and address avenues for future research in section~\ref{sec:conclusion}.

\section{Background}\label{sec:bg}
\subsection{PDDL Planning}
In this paper, we focus on classical, deterministic, fully observable planning tasks specified in the Planning Domain Definition Language (PDDL) \cite{ghallab1998pddl}. For illustration, we refer to the trading domain, where agents navigate a map, trade resources, and deposit them in inventories.

\begin{definition}[PDDL Planning Task]
    A PDDL planning task $\Pi = \left(\mathcal{D}, \mathcal{P}\right)$, consists of a domain $\mathcal{D}$ and problem $\mathcal{P}$. The domain $\mathcal{D} = (T, P, A)$, defines how the world works and is composed of a set of object \textit{types} $T$, \textit{predicates} $P$ and \textit{actions} $A$. The problem $\mathcal{P} = (O, s_0, S_g)$ contains the task specific information, including the \textit{object} instances $O$, the \textit{initial} state $s_0$ and the set of \textit{goal} states $S_G \subseteq S$.
\end{definition}

A \textit{type} $t = \{o_1, o_2,\dots,o_n\}$, is a set of objects sharing common properties. A \textit{predicate} $p : t_1 \times t_2\times\dots\times t_m \rightarrow \{\text{True}, \text{False}\}$, is a boolean function with object arguments and typing and arity requirements. For example, \texttt{located(?p - person ?l - location)} is a predicate with arity 2, where the first argument is a \texttt{person} and the second a \texttt{location}. A ground atom $p(o_1, o_2, \dots, o_n)$, is a predicate applied to specific object instances: for example \texttt{located(p1 l1)} indicates that person \texttt{p1} is \texttt{located} at location \texttt{l1}.

Planning can be expressed as a search problem: $\Phi = (\Sigma, s_0, S_g)$, where $\Sigma = (S, A, \gamma)$ is the state transition system with set of states $S$, actions $A$ and state transition function $\gamma : S \times A \rightarrow S$ defining which actions can be legally applied in which state.

Each \textit{state} $s = \bigwedge_{i=1}^k p_i(o_1, o_2, \dots, o_n)$, is a conjunction of true ground atoms. Actions have preconditions $pre(a)$, add effects $eff^+(a)$, delete effects $eff^-(a)$, and may have an associated cost $c(a)$. The default cost is $1$. An action $a$ is applicable in $s$ if $pre(a) \subseteq s$, such that $\gamma(s,a) = s'$ yields a new state $s' = (s \setminus eff^-(a)) \cup eff^+(a)$. See Listing~\ref{fig:action} for an example from the trading domain. An example action instance could be: \texttt{(deposit p1 r1 l1 i1)}. This action would be applicable in a state if the following preconditions hold: the person \texttt{p1} is \texttt{located} at the location \texttt{l1}, \texttt{l1} \texttt{contains} the inventory \texttt{i1} and \texttt{p1} is \texttt{carrying} the resource \texttt{r1}. In the resulting state, \texttt{p1} will no longer be \texttt{carrying} \texttt{r1} and \texttt{r1} will be \texttt{deposited} in \texttt{i1}.

\begin{lstlisting}[caption={Example action description}, label={fig:action}, style={pddlstyle}, captionpos=b]
(:action deposit
    :parameters (?p - person ?r - resource ?l - location ?i - inventory)
    :precondition (and (located ?p ?l) (contain ?i ?l)
        (carrying ?p ?r))
    :effect (and (not (carrying ?p ?r)) (deposited ?r ?i)))
\end{lstlisting}

The goal of planning is to find a sequence of actions, or plan $\pi = \langle a_1, a_2,\dots,a_n\rangle$. A plan is \textit{valid} if, the actions applied sequentially to the initial state $s_0$, reaches a goal state: $\gamma(\dots\gamma(\gamma(s_0, a_1), a_2),\ldots,a_n) \in S_g$.
The cost of a plan is denoted as $c(\pi) = \sum_{i=1}^n c(a_i)$.

\subsection{Generalized Planning}
Generalized planning refers to the process of generating strategies that are valid for multiple problem instances \cite{martin2004learning}:

\begin{definition}[Generalized Planning Instance]
    A generalized planning instance is a finite set of classical planning tasks $\Pi_G = \{\Pi_1, \Pi_2, \dots, \Pi_n\}$, sharing a common structure.
\end{definition}

In line with recent definitions \cite{jimenez2019review}, we restrict our analysis to the case where the tasks have the same PDDL domain, such that only the objects, initial state and goals are different. A generalized plan, is a solution to a generalized planning instance:

\begin{definition}[Generalized Plan]
    A generalized plan $\Phi$ for $\Pi_G$ is a function that maps all problem instances $\Pi\in \Pi_G$ to valid plans $\pi$.
\end{definition}

Note, to avoid confusing the term \textit{generalized plan} (the method which outputs a plan for a given planning task) and \textit{plan} (the solution to the planning task); we henceforth refer to a generalized plan as a \textit{planner}. 

\subsection{Generalized Planning as an Optimization Problem}
In this paper we are interested in finding \textit{high quality} plans. We can do this by treating the problem as an \textit{optimization problem}, where we seek to find the planner $\Phi$, minimizing plan cost $c(\pi)$, across all tasks $\Pi \in \Pi_G$.
\begin{definition}[Generalized Planning Optimization Problem]
A generalized planning optimization problem is an optimization problem:
\begin{equation}
    \arg\min_{\Phi} \frac{1}{|\Pi_G|}\sum_{\Pi\in\Pi_G} c(\Phi(\Pi))
    \label{eq:fitness}
\end{equation}
\label{def:gpop}
\end{definition}
We consider only the problem of minimizing the number of actions in the plan, such that plan cost $c(\pi) = |\pi|$, referred to as \textit{plan length}, hence the overall objective function is just the mean plan length across all tasks.

\subsection{Evolutionary Algorithms}
Given an optimization problem: $\arg\min_{x} f(x)$, where $\mathcal{X}$ is the feasible solution space and $f: \mathcal{X} \rightarrow \mathbb{R}$ is the objective function to be minimized, the goal is to find $x^* \in \mathcal{X}$ such that $f(x^*) \leq f(x)$ for all $x \in \mathcal{X}$. Evolutionary algorithms are meta-heuristic optimization methods which can find good approximate solutions to complex optimization problems by evolving a population of candidates through selection, crossover, mutation, and replacement.

\begin{definition}[Population]
    Evolutionary algorithms maintain a \textit{population} of candidate solutions: $\mathcal{P}(t) = \{x_1^t, x_2^t, \ldots, x_{\mu}^t\}, \; \text{where } x_i^t \in \mathcal{X}$. The population $\mathcal{P}(t)$ at generation $t$ consists of $\mu$ individual candidates, such that each candidate $x_i^t$ is a feasible solution to the optimization problem: $\min_{x\in \mathcal{X}} f(x)$.\label{def:population}
\end{definition}

Candidates are evaluated using a \textit{fitness function}, which quantifies the quality of the candidate. This guides the evolutionary process by determining which solutions are more likely to survive and reproduce. 

\begin{definition}[Fitness Function]
    The fitness function $\hat{f}:\mathcal{X} \rightarrow \mathbb{R}$, maps each candidate solution to a value measuring how well the solution solves the optimization problem.\label{def:fitness}
\end{definition}

On each iteration, new candidates are identified by applying three operators: \textit{selection}, \textit{crossover} and \textit{mutation}. Parents for crossover are chosen via the \textit{selection} operator.

\begin{definition}[Selection]
    A \textit{selection} operator $\tilde{S}\;:\; \mathcal{X}^{\mu} \rightarrow \mathcal{X}^{k}$, chooses individuals from the current population to become parents for producing offspring. Given the current population $\mathcal{P}(t)$, the selection operator produces a parent pool $\mathcal{P}_{\text{parents}}(t) = \tilde{S}(\mathcal{P}(t))$.\label{def:selection}
\end{definition}

Selection is directly based on the fitness function values $\hat{f}(x)$, with better-fit individuals having higher probability of being selected for reproduction. The crossover operator combines multiple parents to generate offspring.

\begin{definition}[Crossover]
    A \textit{crossover} operator $C\;:\; \mathcal{X}^k \rightarrow \mathcal{X}^\lambda$, takes $k$ selected parents ${s_1, s_2, \ldots, s_k}$, and generates $\lambda$ offspring ${y_1, \ldots, y_\lambda} = C(x_{s_1}^t, \ldots, x_{s_k}^t)$ where each $x_{s_j}^t \in \mathcal{P}_{\text{parents}}(t)$.\label{def:crossover}
\end{definition}

The mutation operator introduces random changes to maintain diversity. Mutation is typically applied with a small probability $p_m \ll 1$, to ensure that genetic material is mostly preserved while still allowing for occasional exploration of new solution characteristics.

\begin{definition}[Mutation]
    A \textit{mutation} operator $M\;:\; \mathcal{X} \rightarrow \mathcal{X}$, introduces random variations to a candidate solution. For each offspring $y_j$ produced by crossover, mutation can be applied to create a modified offspring $y'_j = M(y_j)$.\label{def:mutation}
\end{definition}
After selection, crossover, and mutation, $\lambda$ offspring are produced and added to the cumulative offspring population, $\mathcal{P}'(t)$. This process repeats until reaching a threshold number of offspring $\lambda_{\text{max}}$. The next generation $\mathcal{P}(t+1)$ is then formed by selecting $\mu$ individuals from $\mathcal{P}(t) \cup \mathcal{P}'(t)$ using a replacement strategy~\cite{smith2007replacement}.


\begin{definition}[Replacement Strategy] A \textit{replacement strategy} $R\;:\; \mathcal{X}^{\mu} \times \mathcal{X}^{\lambda} \rightarrow \mathcal{X}^{\mu}$, constructs the next generation population $\mathcal{P}(t+1)$ by selecting exactly $\mu$ individuals from the current population $\mathcal{P}(t)$ and/or the offspring population $\mathcal{P}'(t)$. Formally: $\mathcal{P}(t+1) = R(\mathcal{P}(t), \mathcal{P}'(t)) = \{x_1^{t+1}, x_2^{t+1}, \ldots, x_{\mu}^{t+1}\}$.
\end{definition}

At any generation $t$, the incumbent solution $x^*_t$, is the best solution found so far: $x^*_{t} = \arg\min_{x\in \{\mathcal{P}(t)\cup \mathcal{P}'(t)\}} \hat{f}(x)$. The process repeats for $G$ generations, after which the best solution $x^*_{G}$ is returned as an approximation to the global optimum $x^*$. Evolutionary algorithms do not guarantee optimality, but often yield high-quality solutions when exact optimization is infeasible. See Algorithm~\ref{alg:evo} for pseudocode.


\begin{algorithm}[t]\caption{Evolutionary Algorithm Pseudocode}\footnotesize\begin{enumerate}
    \item Initialize population $\mathcal{P}(0) = {x_1^0, x_2^0, \ldots, x_{\mu}^0}$
    \item Evaluate fitness $\hat{f}(x_i^0)$ for each individual in $\mathcal{P}(0)$
    \item Set $t \leftarrow 0$ \item While $t \leq G$:
    \begin{enumerate}
    \item Initialize offspring population $\mathcal{P}'(t) \leftarrow \emptyset$
    \item While $|\mathcal{P}'(t)| < \lambda_{\text{max}}$
        \begin{enumerate}
            \item Select parents $\mathcal{P}_{\text{parents}}(t)$ based on fitness values
            \item Apply crossover to produce offspring ${y_1, \ldots, y_\lambda}$
            \item Apply mutation to offspring to produce ${y'_1, \ldots, y'_\lambda}$
            \item Evaluate fitness of offspring: $\hat{f}(y'_j)$ for each $j \in {1, \ldots, \lambda}$
            \item $\mathcal{P}'(t) \leftarrow \mathcal{P}'(t) \cup \{y'_1, \ldots, y'_\lambda\}$
        \end{enumerate}
        \item Apply replacement strategy to form new population $\mathcal{P}(t+1)$
        \item $t \leftarrow t + 1$
    \end{enumerate}
    \item Return best solution found $x^*_{G} = \arg\min_{x\in \mathcal{P}(G)} \hat{f}(x)$
\end{enumerate}\label{alg:evo}\end{algorithm}

\section{Related Work}\label{sec:rw}
In this section we discuss relevant related work on planning using LLMs. For a comprehensive overview we refer the reader to a relevant survey \cite{pallagani2024prospects,chiari2025planning}.

\paragraph{Hybrid Approaches} Early studies found LLMs had limited success generating plans directly from PDDL~\cite{silver2022pddl,valmeekam2023planning}. Hybrid approaches overcome this by using LLMs to translate natural language to PDDL which is then solved by downstream planners~\cite{liu2023llm+,dagan2023dynamic,zhang2025lamma, mahdavi2024leveraging, guan2023leveraging}. These approaches often require expert intervention to ensure the generated PDDL is correct and unambiguous. Errors or ambiguities in translation can result in unsolvable or incorrect plans, and the process may fail to capture the full range of user intent due to the inherent ambiguity of natural language.

\paragraph{LLM + Search} Several works used LLMs to perform specific search functions: for example generating successor and goal test functions \cite{katz2025thought}, action selection \cite{stein2025automating} and heuristic generation \cite{meng2024llm,correa2025classical}. These approaches all rely on conventional, often costly, search procedures. In contrast, GenePlan directly generates efficient, interpretable Python implementations, eliminating the need for integration with existing search algorithms.

\paragraph{Instruction Tuning} \citet{verma2025teaching} fine-tune an LLM on labeled planning tasks with explicit reasoning about action preconditions, effects, and state transitions. They enhance performance through step-by-step CoT reasoning with external validation. However, this requires direct model weight access for retraining, while GenePlan works with API-only access, making it more broadly accessible.

\paragraph{Generalized Planning} In many planning domains, simple strategies exist via a generalized plan~\cite{martin2004learning}. Leveraging these strategies can be more efficient than relying on search algorithms, as they provide reusable solutions that generalize across problem instances. Generalized planning has been studied without generative AI, e.g., by learning abstract actions~\cite{bonet2019learning} or policies via deep RL~\cite{rivlin2020generalized}; see~\cite{jimenez2019review} for a survey. \citet{winner2003distill,winner2007loopdistill} learn conditional and looping strategies by exploiting the underlying problem structure.  These approaches learn simple strategies without accounting for any domain specific heuristics that could improve the plan quality.

\citet{wang2024planning} use LLMs to generate planners by creating natural language hints and strategies, converting them into solution sketches, then translating to pseudocode and finally Python code. This approach is not designed for PDDL planning problems. \citet{silver2022pddl,silver2024generalized} showed that chain-of-thought prompting can extract Python strategies for generalized planning. A similar approach is taken by \citet{chen2025language} using few-shot-prompting. These approaches are satisficing and are not optimized. Our work addresses this gap by focusing on plan quality optimization.

\paragraph{Evolutionary LLMs} Recent work has demonstrated that evolutionary frameworks leveraging LLMs can automatically generate and refine Python methods for combinatorial optimization~\cite{romera2024mathematical,liu2024evolution}. By iteratively applying selection, crossover, and mutation to candidate heuristics, these approaches have produced solutions that surpass the best human-designed heuristics. \citet{liu2024evolution} further introduced prompt strategies inspired by evolutionary algorithms to enhance exploration and diversity in generated solutions. This approach has been successfully applied to other tasks, for example automated feature engineering \cite{abhyankar2025llm,murray2025elate,gong2025evolutionary}, drug discovery~\cite{wang2024efficient} and generating financial trading strategies \cite{yuksel2025alphaquant,yuksel2025evorisk}. Our work extends this paradigm to generalized PDDL planning.

\section{Method}\label{sec:method}
Our insight is to frame the Generalized Planning Optimization Problem (Definition~\ref{def:gpop}) as searching for a generalized planner $\Phi$ within the feasible set $\mathcal{X}$ of executable code. In our experiments, $\mathcal{X}$ is restricted to valid Python code, though the approach is language-agnostic. We introduce GenePlan, an evolutionary algorithm that evolves a population of planners to approximately minimize Equation~\eqref{eq:fitness}.

\begin{figure}
    \centering
    \includegraphics[width=\columnwidth]{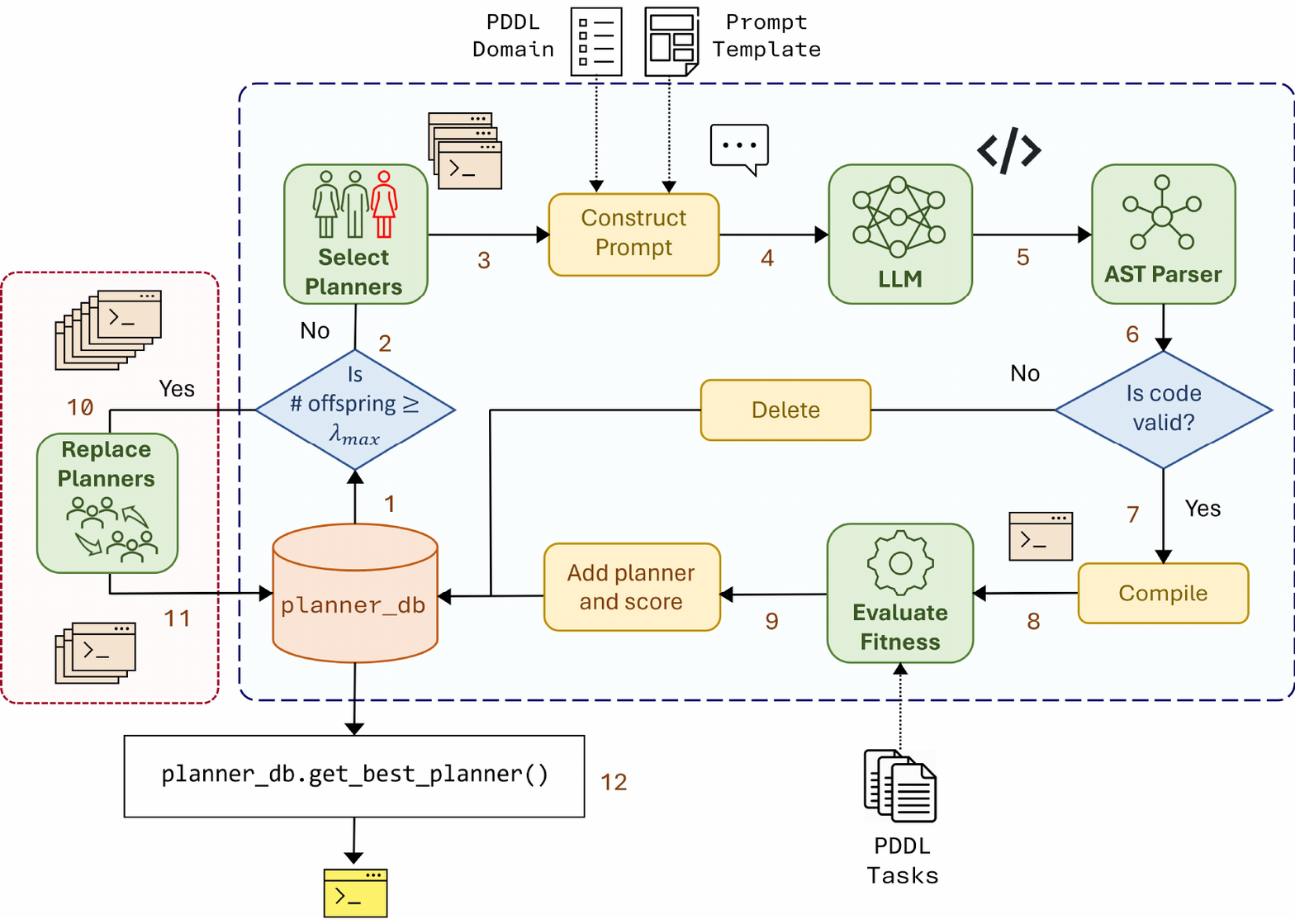}
    \caption{Figure showing architecture of GenePlan. Python planners  are stored in the \texttt{planner\_db}. The \textcolor{blue}{right} loop (1-9) generates new candidate planners, while the \textcolor{red}{left} loop (10-11) prunes low scoring candidates at the end of each generation. After running, the best planner can be extracted by querying the \texttt{planner\_db} (12)}
    \label{fig:GenePlan}
\end{figure}

The GenePlan architecture is shown in Figure \ref{fig:GenePlan} and is inspired by \citet{romera2024mathematical}. It can be thought of as an evolutionary algorithm, whereby the right loop (1-9) creates candidate offspring planners (stored in the \texttt{planner\_db}) via selection and LLM-assisted recombination and mutation; and the left loop (10-11) prunes the worst performing planners via replacement. After $G$ generations, the best planner is extracted from the \texttt{planner\_db}. Inputs include a prompt template, the PDDL domain $\mathcal{D}$, and training tasks $\Pi_{train} \subseteq \Pi_G$. The initial population can be seeded with a provided Python planner or generated via chain-of-thought prompting \cite{silver2024generalized}. Details of each component are described below.

\paragraph{Evaluate Fitness}
Each candidate is stored alongside its fitness function score within the \texttt{planner\_db}. The fitness function is computed in the \textit{evaluate fitness}, block 8 in Figure \ref{fig:GenePlan}. As a surrogate of Equation \eqref{eq:fitness}, our fitness function simply computes the average plan length of the plans found by the planner on the set of training tasks:
\begin{equation}
    \hat{f}(\Phi) = \frac{1}{|\Pi_{train}|}\sum_{\Pi\in \Pi_{train}}|\Phi(\Pi)|
\end{equation}
Note that each plan output from the planner is first validated using a plan validator \cite{howey2003val} to ensure validity. We define a failure score $F \gg 0$, such that if a planner $\Phi$ us unable to solve a particular instance $\Pi \in \Pi_{train}$, then the score for that problem $|\Phi(\Pi)| = F$.

\paragraph{Select Planners}
Parent planners are selected in the \textit{select planners} part (2) in Figure \ref{fig:GenePlan}. As per~\citet{romera2024mathematical}, probabilities are assigned to each planner based on their fitness scores. Unlike \citeauthor{romera2024mathematical}, we chose to use an inverse decaying function: $T = a/|\mathcal{P}(t) \cup \mathcal{P}'(t)| + b$ to model the temperature $T$, at any iteration as a function of the current number of planners in the \texttt{planner\_db}, $|\mathcal{P}(t) \cup \mathcal{P}'(t)|$, where $b = (T_{\text{min}}(\mu + \lambda_{\text{max}}) - T_{\text{max}})/(\mu+\lambda_{\text{max}}-1)$ and $a = T_{\text{max}} - b$. The temperature $T$ decays hyperbolically from $T_{\text{max}}$ to $T_{\text{min}}$ as the number of planners in the \texttt{planner\_db} increases from 1 to $\mu + \lambda_{\text{max}}$.  We use $T_{\text{max}} = 50$, $T_{\text{min}} = 10$ and $\mu = 10$ and $\lambda_{\text{max}} = 10$ in our experiments.

Given the temperature $T$ at any given iteration, we can compute the probability of sampling a planner $\Phi_j$ as:
\begin{equation} P_{\Phi_j} = \frac{e^{\left(-\hat{f}(\Phi_j) / T\right)}}{\sum_{\Phi \in {\mathcal{P}(t) \cup \mathcal{P}'(t)}}e^{\left(-\hat{f}(\Phi) / T\right)}}\label{eq:select} \end{equation}

The lower the temperature, the higher probability is assigned to planners with low values of $\hat{f}$. Since the temperature is hyperbolically decaying, this has the effect of encouraging exploration early on in the generation, and exploitation towards the end.  Equation \eqref{eq:select} defines a discrete probability distribution $P$ over the set of planners ${\mathcal{P}(t) \cup \mathcal{P}'(t)}$, where each planner $\Phi_j$ has probability $P_{\Phi_j}$ of being sampled. The distribution satisfies $\sum_{\Phi \in {\mathcal{P}(t) \cup \mathcal{P}'(t)}} P_{\Phi} = 1$ and $P_{\Phi} \geq 0$ for all $\Phi$.

The selection operator $\tilde{S}$ samples $k$ parent planners without replacement from ${\mathcal{P}(t) \cup \mathcal{P}'(t)}$ according to $P_{\Phi}$, yielding $\mathcal{P}_{\text{parents}}(t) = {\Phi_{s_1}, \Phi_{s_2}, \ldots, \Phi_{s_k}}$. Note that we sample from both the current population and cumulative offspring, unlike Definition \ref{def:population}. If fewer than $k$ planners are available, we sample $\min(|\mathcal{P}(t) \cup \mathcal{P}'(t)|, k)$; crossover is skipped when only one planner can be sampled.

\paragraph{Prompt Construction}
The prompt template used in the experiments is provided in Listing \ref{fig:prompt-template}. The placeholder \texttt{@@domain@@} is replaced with the PDDL domain. Parent planners sampled in the selection phase are inserted in place of \texttt{@@examples@@} to construct an n-shot prompt (3). Note that for each planner, if the code fails, we catch the exception and pass the error message \texttt{planner.error}. If the planner fails to solve some problems, then we pass the result of the plan validator \texttt{planner.plan\_failure\_error} for these instances. This prompt is then passed to the LLM (4) and a new offspring candidate $y'$ is generated. Note that this step encompasses both the evolutionary \textit{crossover} and \textit{mutation} operators.
\begin{figure}[t!]
\begin{lstlisting}[style=textsmall, 
                     linerange={1-6}]
You have access to the following PDDL domain:

@@domain@@

Your goal is to implement a method called get_plan in Python that solves problem instances for this domain with the fewest possible actions. The method should be of the form:
\end{lstlisting}
    
\begin{lstlisting}[style=pythonstylesmall, 
                     linerange={7-12}]
def get_plan(objects, init, goal):
    """
    Description of heuristic.
    """
    Your code here
    return plan
\end{lstlisting}
    
\begin{lstlisting}[style=textsmall, 
                     linerange={13-34}]
where
- `objects` is a @@typing@@.
- `init` is a set of ground atoms represented as tuples of predicate names and 1 or more arguments (e.g., ('predicate-foo', 'object-bar', ...))
- `goal` is also a set of ground atoms represented in the same way
- `plan` is a list of actions, where each action is a ground operator represented as a string (e.g., '(operator-baz object-qux ...)')

I will now provide you with some example methods. Below each method you will be provided with a system message telling you whether the code worked or not. This will be either:

System: The code did not work. Error: {planner.error}.
System: The code failed to solve some problems. Error: {planner.plan_failure_error}. Score {planner.score}
System: The code worked. Score: {planner.score}.

The score is the average plan length across all test problems (lower is better). Here are the examples:

@@examples@@

EVOLUTIONARY OPTIMIZATION INSTRUCTIONS:

Given these examples, your task is to use evolutionary optimization to generate a better heuristic:
- Perform crossover: Combine the best algorithmic components from each method. Merge these segments in a way that preserves their functionality.
- Apply mutations: Introduce strategic modifications such as efficiency improvements, novel heuristic calculations, redundant action elimination, or alternative goal prioritization.
- Optimize for fitness: Optimize to minimize the average plan length (score). Consider both solution quality and computational efficiency.
\end{lstlisting}
    
    \caption{Evolutionary prompt template.}
    \label{fig:prompt-template}
\end{figure}

\paragraph{AST Parser}
At (5), the LLM outputs Python code as a string. Prior to execution we validate it using an abstract syntax tree (AST) parser. This ensure it adheres to a predefined set of allowable nodes, packages, functions, and attributes, serving as a guardrail. If the code is valid (6), it is compiled and executed to produce a method, stored with the code string as a planner object. The fitness of valid offspring planners is evaluated (8) and then added to the \texttt{planner\_db} (9). Invalid planners are deleted, returning to (1).

\paragraph{Replace Planners}
This process repeats (1 - 9), until the number of offspring planners stored in the \texttt{planner\_db} reaches the maximum $\lambda_{\text{max}}$. At this point replacement occurs in the left loop (10-11), whereby we replace the current population with the best performing candidates from the generation. As a replacement strategy, we initialize the next generation's population using the planners with the lowest (best) fitness score:
$
\mathcal{P}(t+1) = \{\Phi \in \mathcal{P}(t)\cup\mathcal{P}'(t)\;\mid\;\text{rank}_{\hat{f}}(\Phi)\leq \mu\}
$, where $\text{rank}_{\hat{f}}(\Phi)$ gives the rank of individual $\Phi$ when all individuals in $\mathcal{P}(t) \cup \mathcal{P}'(t)$ are sorted by their fitness values $\hat{f}$ in ascending order: $\text{rank}_{\hat{f}}(\Phi) = 1 + |\{\Phi_j \in \mathcal{P}(t) \cup \mathcal{P}'(t)\;\mid \;\hat{f}(\Phi_j) < \hat{f}(\Phi)\}|$. Note that we select the next generation from both the previous generation $\mathcal{P}(t)$, and the current offspring $\mathcal{P}'(t)$. This is referred to as $\mu + \lambda$ selection ($\mu$ members in the population + $\lambda$ offspring) or an \textit{elitist} replacement strategy \cite{beyer2002evolution}.

\section{Experimental Setup}\label{sec:experiments}
We experimentally validated our method on 8 PDDL domains. Separate train (for use within GenePlan) and test (for evaluation) problems were generated for each domain. We used 30 test problems for each domain and 5-10 training tasks. A summary of the domains and their source are provided below. Most of these domains have been extracted from prior studies \cite{yang2022pg3,silver2024generalized}. To ensure that all of the domains are not within the LLMs training set, we also created two entirely new domains (trading and research). All domains contain problems of varying sizes and initial and goal configurations. We include below, details of problem sizes for the new domains \footnotemark.  Full PDDL for these domains is provided in Appendix \ref{app:domains}

\begin{itemize}
\item \textbf{Trapnewspapers} \cite{yang2022pg3}: Pick up newspapers from a base and deliver them to specified locations.
\item \textbf{Hiking} \cite{yang2022pg3}: Navigate a map, climbing hills and avoiding water obstacles.
\item \textbf{Manygripper} \cite{yang2022pg3}: Use a robot with two grippers to transport balls between locations.
\item \textbf{ManyFerry} \cite{yang2022pg3}: Transport cars between islands using ferries.
\item \textbf{Manymiconic}\cite{yang2022pg3}: Pick up and drop off passengers in multi-building elevator systems.
\item \textbf{Heavypack}\cite{silver2024generalized}: Stack items in a box, only allowing heavier items to be stacked upon.
\item \textbf{Research}: Agents read papers, run experiments, write papers, and advisors can teach to accelerate learning.
\item \textbf{Trading}: Agents move around a map, gather resources, deposit them in inventories, and trade resources with other agents at the same location.
\end{itemize}
\footnotetext{Problem sizes for new domains—Research: train (5-10 researchers, 2-5 projects, 10-20 papers/experiments), test (10-20 researchers, 5-15 projects, 20-50 papers/experiments); Trading: train (2-5 agents/inventories, 10-20 locations/resources), test (5-10 agents/inventories, 20-30 locations/resources).}
\noindent We compare against the following baselines:
\begin{itemize}
\item \textbf{cot4}~\cite{silver2024generalized}: Chain-of-thought prompting with GPT-4.

\item \textbf{cot4o}: Chain-of-thought prompting with GPT-4o.

\item \textbf{evo}: The evolutionary planner proposed in this paper, using GPT-4o.

\item \textbf{evo\_ds}: Evo variant where the LLM generates a natural language domain summary for the prompt instead of using the PDDL domain.

\item \textbf{evo\_mini}: Evo variant using GPT-4o mini.

\item \textbf{evo\_abl}: Evo variant with predicate, object and action names ablated. This was done by replacing names with generic names like \texttt{action1} and \texttt{predicate1}.

\item \textbf{evo\_nev}: Evo variant with no evaluator; all planners receive constant scores.

\item \textbf{fd\_$x$}:  \textsc{lama} setting~\cite{richter2010lama} of Fast Downward~\cite{helmert2006fast}, which is an anytime portfolio approach designed to find higher-quality plans within a specified time bound $x$. We evaluate two configurations: $x=300$s and $x=1800$s.

\item \textbf{fd\_opt}: \textsc{seq-opt-lmcut} setting of Fast Downward, which runs $A^*$ with the admissible \textsc{lmcut} heuristic~\cite{helmert2009landmarks} to compute an optimal plan. We let this configuration run for 30 minutes.
\end{itemize}

We used the unified planning framework \cite{micheli2025unified} for all experiments.

\section{Results}\label{sec:results}
 Results are provided in Table \ref{tab:performance_comparison}.

\begin{table*}[t]
\centering
\scriptsize
\begin{tabular}{llcccccccc}
\toprule
 & domain & heavypack & hiking & manyferry & manygripper & manymiconic & research & trading & trapnewspapers \\
method & metric &  &  &  &  &  &  &  &  \\
\midrule
\multirow[t]{3}{*}{fd\_300} & \% Solved & \textbf{100.0} & \textbf{100.0} & \textbf{100.0} & \textbf{100.0} & \textbf{100.0} & 90.0 & 43.33 & 83.33 \\
 & Runtime & 21.19 {\tiny $\pm$ 2.05} & 5.45 {\tiny $\pm$ 0.27} & 18.69 {\tiny $\pm$ 4.20} & 63.57 {\tiny $\pm$ 13.28} & 22.86 {\tiny $\pm$ 5.55} & 83.48 {\tiny $\pm$ 12.57} & 219.94 {\tiny $\pm$ 15.79} & 71.69 {\tiny $\pm$ 16.64} \\
 & Score & \textbf{1.00 {\tiny $\pm$ 0.00}} & 0.89 {\tiny $\pm$ 0.05} & 0.93 {\tiny $\pm$ 0.01} & 0.94 {\tiny $\pm$ 0.01} & 0.98 {\tiny $\pm$ 0.00} & 0.90 {\tiny $\pm$ 0.06} & 0.43 {\tiny $\pm$ 0.09} & 0.64 {\tiny $\pm$ 0.05} \\
\cline{1-10}
\multirow[t]{3}{*}{fd\_1800} & \% Solved & \textbf{100.0} & \textbf{100.0} & \textbf{100.0} & \textbf{100.0} & \textbf{100.0} & \textbf{100.0} & 93.33 & 86.67 \\
 & Runtime & 10.04 {\tiny $\pm$ 0.89} & 2.73 {\tiny $\pm$ 0.09} & 110.76 {\tiny $\pm$ 49.71} & 88.17 {\tiny $\pm$ 27.73} & 49.12 {\tiny $\pm$ 18.65} & 31.87 {\tiny $\pm$ 4.28} & 394.48 {\tiny $\pm$ 50.64} & 216.66 {\tiny $\pm$ 30.44} \\
 & Score & \textbf{1.00 {\tiny $\pm$ 0.00}} & 0.89 {\tiny $\pm$ 0.05} & 0.95 {\tiny $\pm$ 0.01} & \textbf{0.96 {\tiny $\pm$ 0.01}} & \textbf{0.99 {\tiny $\pm$ 0.00}} & \textbf{1.00 {\tiny $\pm$ 0.00}} & \textbf{0.93 {\tiny $\pm$ 0.05}} & 0.70 {\tiny $\pm$ 0.05} \\
\cline{1-10}
\multirow[t]{3}{*}{fd\_opt} & \% Solved & \textbf{100.0} & \textbf{100.0} & 56.67 & 26.67 & 53.33 & 0.0 & 0.0 & 0.0 \\
 & Runtime & 11.49 {\tiny $\pm$ 1.37} & 12.68 {\tiny $\pm$ 0.87} & 744.50 {\tiny $\pm$ 143.88} & 953.29 {\tiny $\pm$ 207.66} & 87.54 {\tiny $\pm$ 28.83} & - & - & - \\
 & Score & \textbf{1.00 {\tiny $\pm$ 0.00}} & \textbf{1.00 {\tiny $\pm$ 0.00}} & 0.57 {\tiny $\pm$ 0.09} & 0.27 {\tiny $\pm$ 0.08} & 0.53 {\tiny $\pm$ 0.09} & 0.00 {\tiny $\pm$ 0.00} & 0.00 {\tiny $\pm$ 0.00} & 0.00 {\tiny $\pm$ 0.00} \\
\cline{1-10}
\multirow[t]{3}{*}{cot4} & \% Solved & \textbf{100.0} & \textbf{100.0} & \textbf{100.0} & \textbf{100.0} & 0.0 & \textbf{100.0} & \textbf{100.0} & \textbf{100.0} \\
 & Runtime & 1.15 {\tiny $\pm$ 0.09} & 0.91 {\tiny $\pm$ 0.02} & \textbf{0.18 {\tiny $\pm$ 0.01}} & \textbf{0.45 {\tiny $\pm$ 0.01}} & - & 5.95 {\tiny $\pm$ 0.51} & 0.24 {\tiny $\pm$ 0.01} & 0.10 {\tiny $\pm$ 0.00} \\
 & Score & \textbf{1.00 {\tiny $\pm$ 0.00}} & 0.55 {\tiny $\pm$ 0.06} & 0.90 {\tiny $\pm$ 0.01} & 0.90 {\tiny $\pm$ 0.01} & 0.00 {\tiny $\pm$ 0.00} & 0.06 {\tiny $\pm$ 0.00} & 0.21 {\tiny $\pm$ 0.01} & 0.75 {\tiny $\pm$ 0.00} \\
\cline{1-10}
\multirow[t]{3}{*}{cot4o} & \% Solved & \textbf{100.0} & \textbf{100.0} & \textbf{100.0} & \textbf{100.0} & 0.0 & \textbf{100.0} & \textbf{100.0} & \textbf{100.0} \\
 & Runtime & 1.25 {\tiny $\pm$ 0.10} & 0.90 {\tiny $\pm$ 0.02} & 0.21 {\tiny $\pm$ 0.01} & 0.50 {\tiny $\pm$ 0.01} & - & 0.93 {\tiny $\pm$ 0.05} & 0.15 {\tiny $\pm$ 0.00} & 0.12 {\tiny $\pm$ 0.00} \\
 & Score & \textbf{1.00 {\tiny $\pm$ 0.00}} & 0.55 {\tiny $\pm$ 0.06} & 0.92 {\tiny $\pm$ 0.01} & 0.90 {\tiny $\pm$ 0.01} & 0.00 {\tiny $\pm$ 0.00} & 0.11 {\tiny $\pm$ 0.01} & 0.88 {\tiny $\pm$ 0.01} & 0.75 {\tiny $\pm$ 0.00} \\
\cline{1-10}
\multirow[t]{5}{*}{evo\_nev} & \% Solved & \textbf{100.0} & \textbf{100.0} & \textbf{100.0} & \textbf{100.0} & 16.67 & \textbf{100.0} & \textbf{100.0} & \textbf{100.0} \\
 & Runtime & 1.24 {\tiny $\pm$ 0.10} & 0.90 {\tiny $\pm$ 0.02} & 0.24 {\tiny $\pm$ 0.02} & 0.47 {\tiny $\pm$ 0.01} & 0.23 {\tiny $\pm$ 0.05} & 5.88 {\tiny $\pm$ 0.50} & 0.25 {\tiny $\pm$ 0.01} & 0.10 {\tiny $\pm$ 0.00} \\
 & Score & \textbf{1.00 {\tiny $\pm$ 0.00}} & 0.55 {\tiny $\pm$ 0.06} & 0.90 {\tiny $\pm$ 0.01} & 0.90 {\tiny $\pm$ 0.01} & 0.11 {\tiny $\pm$ 0.05} & 0.06 {\tiny $\pm$ 0.00} & 0.19 {\tiny $\pm$ 0.01} & 0.75 {\tiny $\pm$ 0.00} \\
 & Cost & 1.26 & 1.45 & 1.65 & 1.66 & 1.94 & 2.32 & 1.9 & 1.22 \\
 & Gen Time & 460.61 & 602.82 & 620.1 & 718.18 & 13359.07 & 853.06 & 901.03 & 462.0 \\
\cline{1-10}
\multirow[t]{5}{*}{evo\_abl} & \% Solved & 0.0 & 0.0 & 0.0 & 0.0 & 0.0 & 0.0 & 0.0 & 0.0 \\
 & Runtime & - & - & - & - & - & - & - & - \\
 & Score & 0.00 {\tiny $\pm$ 0.00} & 0.00 {\tiny $\pm$ 0.00} & 0.00 {\tiny $\pm$ 0.00} & 0.00 {\tiny $\pm$ 0.00} & 0.00 {\tiny $\pm$ 0.00} & 0.00 {\tiny $\pm$ 0.00} & 0.00 {\tiny $\pm$ 0.00} & 0.00 {\tiny $\pm$ 0.00} \\
 & Cost & - & - & - & - & - & - & - & - \\
 & Gen Time & - & - & - & - & - & - & - & - \\
\cline{1-10}
\multirow[t]{5}{*}{evo\_mini} & \% Solved & \textbf{100.0} & \textbf{100.0} & \textbf{100.0} & \textbf{100.0} & 0.0 & \textbf{100.0} & \textbf{100.0} & \textbf{100.0} \\
 & Runtime & 1.14 {\tiny $\pm$ 0.09} & 1.03 {\tiny $\pm$ 0.06} & 0.19 {\tiny $\pm$ 0.01} & 0.46 {\tiny $\pm$ 0.01} & - & 5.90 {\tiny $\pm$ 0.50} & \textbf{0.14 {\tiny $\pm$ 0.00}} & 0.11 {\tiny $\pm$ 0.00} \\
 & Score & \textbf{1.00 {\tiny $\pm$ 0.00}} & 0.55 {\tiny $\pm$ 0.06} & 0.90 {\tiny $\pm$ 0.01} & 0.91 {\tiny $\pm$ 0.01} & 0.00 {\tiny $\pm$ 0.00} & 0.06 {\tiny $\pm$ 0.00} & 0.92 {\tiny $\pm$ 0.01} & 0.75 {\tiny $\pm$ 0.00} \\
 & Cost & \textbf{0.07} & \textbf{0.09} & \textbf{0.08} & \textbf{0.1} & \textbf{0.12} & \textbf{0.15} & \textbf{0.15} & \textbf{0.08} \\
 & Gen Time & 895.7 & 16105.96 & 1111.67 & 1128.4 & 61870.86 & 1589.72 & 2061.29 & 1088.75 \\
\cline{1-10}
\multirow[t]{5}{*}{evo\_ds} & \% Solved & \textbf{100.0} & \textbf{100.0} & \textbf{100.0} & 73.33 & 0.0 & \textbf{100.0} & \textbf{100.0} & \textbf{100.0} \\
 & Runtime & \textbf{1.12 {\tiny $\pm$ 0.08}} & 0.93 {\tiny $\pm$ 0.02} & 0.19 {\tiny $\pm$ 0.01} & 0.50 {\tiny $\pm$ 0.01} & - & 1.49 {\tiny $\pm$ 0.11} & 0.15 {\tiny $\pm$ 0.00} & \textbf{0.09 {\tiny $\pm$ 0.00}} \\
 & Score & \textbf{1.00 {\tiny $\pm$ 0.00}} & 0.89 {\tiny $\pm$ 0.05} & \textbf{0.98 {\tiny $\pm$ 0.00}} & 0.70 {\tiny $\pm$ 0.08} & 0.00 {\tiny $\pm$ 0.00} & 0.29 {\tiny $\pm$ 0.01} & 0.91 {\tiny $\pm$ 0.01} & \textbf{1.00 {\tiny $\pm$ 0.00}} \\
 & Cost & 1.3 & 1.55 & 1.89 & 1.76 & 2.04 & 2.52 & 1.87 & 1.44 \\
 & Gen Time & \textbf{437.16} & \textbf{573.33} & 721.06 & \textbf{574.7} & \textbf{630.98} & \textbf{738.03} & 659.57 & 1761.47 \\
\cline{1-10}
\multirow[t]{5}{*}{evo} & \% Solved & \textbf{100.0} & \textbf{100.0} & \textbf{100.0} & \textbf{100.0} & \textbf{100.0} & \textbf{100.0} & \textbf{100.0} & \textbf{100.0} \\
 & Runtime & 1.15 {\tiny $\pm$ 0.09} & \textbf{0.88 {\tiny $\pm$ 0.02}} & 0.19 {\tiny $\pm$ 0.01} & 0.49 {\tiny $\pm$ 0.01} & \textbf{0.20 {\tiny $\pm$ 0.02}} & \textbf{0.75 {\tiny $\pm$ 0.04}} & 0.15 {\tiny $\pm$ 0.00} & 0.10 {\tiny $\pm$ 0.00} \\
 & Score & \textbf{1.00 {\tiny $\pm$ 0.00}} & 0.89 {\tiny $\pm$ 0.05} & 0.95 {\tiny $\pm$ 0.01} & 0.90 {\tiny $\pm$ 0.01} & 0.98 {\tiny $\pm$ 0.01} & 0.71 {\tiny $\pm$ 0.02} & 0.92 {\tiny $\pm$ 0.01} & 0.94 {\tiny $\pm$ 0.01} \\
 & Cost & 1.26 & 1.67 & 1.62 & 2.02 & 2.25 & 2.81 & 1.83 & 1.11 \\
 & Gen Time & 461.24 & 657.0 & \textbf{610.51} & 745.6 & 703.94 & 933.97 & \textbf{653.72} & \textbf{396.38} \\
\bottomrule
\end{tabular}

\caption{Performance comparison across different methods and domains}
\label{tab:performance_comparison}
\end{table*}

\paragraph{SAT Score}

To compare planner performance across domains and baselines, we use the SAT score $SAT_m(\Pi) = c(\pi^*)/c(\pi_m)$, from the International Planning Competition (IPC)~\cite{taitler20242023}, where $c(\pi^*)$ is the cost of the best plan $\pi^*$, found across all methods and $c(\pi_m)$ is the cost from the method $m$ under evaluation. A score of $1$ means the method found the best plan, whereas $0$ indicates an unsolved task. Table~\ref{tab:performance_comparison} presents the mean and standard error SAT scores for all methods and domains. Our evolutionary planner (evo) achieved an average SAT score of $0.91$ across all domains, closely matching the performance of Fast Downward with a 30-minute time limit ($0.93$). Notably, evo outperformed other LLM-based baselines with an average improvement in SAT score of $0.27$ versus cot4o ($0.64$).

We observe that using a domain summary (evo\_ds) performed well in simpler domains, such as \textit{manygripper} and \textit{trapnewspapers}, but struggled in more complex domains like \textit{manymiconic}. This suggests that the completeness of domain representation in the prompt can significantly impact plan quality. Consistent with prior studies~\cite{silver2024generalized}, we found that ablating names from the PDDL severely impairs the LLM's ability to generate working solutions. This highlights the importance of contextual information for effective plan synthesis. It's worth mentioning that fd\_opt achieved a much lower score (average of $0.42$) than fd\_1800 with the same time limit. In many domains, fast downward failed to find an optimal plan within 30 minutes resulting in a score of 0 on that problem. In contrast, fd\_1800 returns the best plan found within the time limit.

We would like to highlight that while our experimental analysis used proprietary models (GPT-4o and GPT-4o mini), newer, open source models such as Deepseek-V3 and R1 are available and have shown superior performance on math and coding benchmarks \cite{guo2025deepseek}.

\paragraph{Relative Performance across Problem Instances}
\begin{figure}
    \centering
    \includegraphics[width=0.7\linewidth]{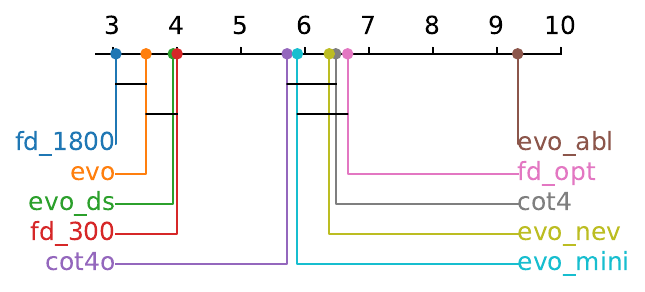}
    \caption{Critical difference diagram showing average rank per method across all problem instances. Lower ranks (left) are better and the horizontal lines connecting approaches indicate statistical indistinguishability.}
    \label{fig:placeholder}
\end{figure}

We performed a comprehensive statistical comparison across all 240 problem instances solved (30 per domain). For each task, methods were ranked by their performance scores, and these ranks were averaged to assess overall effectiveness. To determine whether differences in method performance were statistically significant, we used the Friedman test~\cite{friedman1940comparison}, which compares multiple algorithms across multiple datasets. A significant Friedman test result ($p < 0.05$) indicates that at least one method performs differently. To identify which specific pairs of methods differ, we applied the Nemenyi post-hoc test~\cite{nemenyi1963distribution}, which accounts for the increased risk of false positives when making many pairwise comparisons. This ensures that the chance of incorrectly finding a significant difference remains below a chosen threshold (such as 0.05).

The results are summarized in a critical difference diagram \cite{demvsar2006statistical}, which plots the average ranks of all methods along a horizontal axis, with lower average ranks (better performing methods) positioned further to the left. Horizontal bars connect groups of methods whose differences are not statistically significant at the 0.05 level according to the Nemenyi test. Notably, evo and fd\_$1800$ are grouped together, indicating comparable performance, while both evo and evo\_ds are separated from cot4o, reflecting a statistically significant improvement over the cot4o baseline.

\paragraph{Dollar Cost}
We report the dollar cost of each method, reflecting the resources required to generate solutions using large language models (LLMs). Cost is determined by API usage fees, calculated using OpenAI pricing~\cite{openai_pricing}: \$10/\textit{M} output and \$2.5/\textit{M} input tokens for GPT-4o, and \$0.6/\textit{M} output and \$0.15/\textit{M} input tokens for GPT-4o mini. On average, generating plans with GPT-4o cost just \$1.82 per domain, while GPT-4o mini averaged \$0.10 per domain but with lower plan quality (score 0.64 vs. 0.91 for GPT-4o). This performance gap is primarily due to the smaller size and reduced training data of GPT-4o mini, which hinders its reasoning capabilities on complex tasks.

\paragraph{Runtime}

\begin{table}
\scriptsize
\centering
\caption{Minimum number of instances $k$ required for LLM-based planner to be more efficient than fd\_1800 per domain.}
\label{tab:k_per_domain}
\begin{tabular}{cc}
\toprule
Domain & $k$ \\
\midrule
trading & 1.66 \\
trapnewspapers & 1.83 \\
manyferry & 5.52 \\
manygripper & 8.5 \\
manymiconic & 14.39\\
research & 30.01 \\
heavypack & 51.85 \\
hiking & 354.53 \\
\bottomrule
\end{tabular}
\end{table}

For the evolutionary (evo) methods, we report both the generation time (\textit{Gen Time}) (the total time required to run the evolutionary algorithm and produce a planner—and) the runtime (the average time taken to solve each PDDL problem using the generated planner). In our experiments, evo required on average 645 seconds to generate a planner\_db with the chosen parameters ($\mu = 10, \lambda_\text{max} = 10$, $G = 10$). Importantly, this generation time is a one-time cost; once the planner is produced, it can be reused to solve new instances rapidly, with an average runtime of just 0.49 seconds per plan—significantly faster than Fast Downward.

We define $T_\text{gen}$ as the one-time planner generation time, $T_\text{evo}$ as the average time to solve a single instance with the evolutionary planner, and $T_\text{fd}$ as the average time per instance for Fast Downward (fd\_1800). GenePlan is advantageous when the total time to solve $k$ instances, $T_\text{gen} + k T_\text{evo}$, is less than $k T_\text{fd}$, which occurs for $k > T_\text{gen}/(T_\text{fd} - T_\text{evo})$. Table~\ref{tab:k_per_domain} lists the required $k$ for each domain.

\begin{figure}
    \centering
    \includegraphics[width=\linewidth]{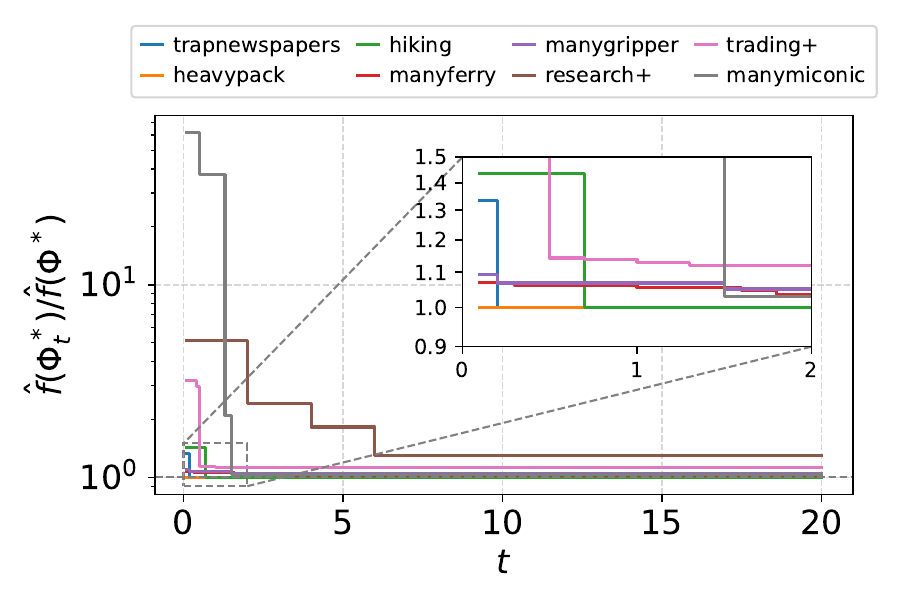}
    \caption{Normalized score $\hat{f}(\Phi^*_t) / \hat{f}(\Phi^*)$ versus generation $t$. $\hat{f}(\Phi^*)$ is the optimal average plan length found by Fast Downward on the training tasks (or best found within 1 hour for domains labelled +), and $\hat{f}(\Phi^*_t)$ is GenePlan's current incumbent solution.}
    \label{fig:runtime}
\end{figure}

The results indicate that for complicated domains like trading, the evolutionary approach becomes advantageous after solving just a few instances. Importantly, our experiments did not use early stopping; for simple domains this would avoid unnecessary computation. As an example, Figure \ref{fig:runtime}  shows the average plan length versus generation on the training tasks, normalized by the optimal value from fast downward. For some domains (trapnewspapers, heavypack and hiking), GenePlan converged on the global optima within a single generation. Future work will explore early stopping criteria, such as halting when improvement plateaus or using an LLM to infer when it believes it has found the optimal solution.

\paragraph{Domains with no Simple Strategy}\label{sec:no_simple_strategy}
Importantly, simple, generalizable strategies are not always available for all planning domains. For example, in Sokoban, certain actions can lead to irreversible states—such as pushing a stone into a corner—making the goal unreachable. We tested GenePlan on this domain and found that it tried to build a search algorithm (see Appendix \ref{app:plan_no_strategy}). The resulting planner failed to solve any problems, while fast downward with a time limit of 30 minutes solved most instances.

In such domains, exhaustive or heuristic search remains necessary. We are not interested in reinventing search based planners which are already highly optimized for this type of problem. An interesting direction would be using an LLM as an orchestration interface to dynamically select between GenePlan-generated and traditional search-based planners based on problem context. This could be facilitated by emerging tools such as the Model Context Protocol (MCP) \cite{hou2025model}. Finally, our approach could be used to generate specific components of search algorithms, for example successor state and goal tests \cite{katz2025thought} or heuristics \cite{correa2025classical}, while leveraging existing search algorithms.

\section{Conclusion}\label{sec:conclusion}
In this paper, we introduced GenePlan, a novel framework that leverages evolutionary LLMs to generate cost-aware, domain-dependent planners for classical PDDL planning tasks. By framing generalized planning as an optimization problem and integrating LLMs within an evolutionary algorithm, our approach evolves interpretable Python planners that minimize plan length across diverse problem instances.

Our empirical evaluation across eight benchmark domains demonstrates that GenePlan achieves plan quality comparable to state-of-the-art planners such as Fast Downward, while offering significant advantages in interpretability and rapid inference. The one-time computational cost of generating a planner is offset by efficiency gains in repeated deployment, making our approach attractive for domains with recurring planning needs. We showed that using a natural language summary of the domain does not generally lead to improved performance, while ablation studies highlight the importance of contextual information for LLM-based planning. Our cost analysis indicates that high-quality planners can be generated at a reasonable expense.

We conclude by listing avenues for future work. The first involves developing early stopping criteria to reduce the computational cost of planner generation. Secondly, new optimization metrics beyond plan length should be investigated. Finally, as mentioned in Section \ref{sec:no_simple_strategy}, LLMs can be leveraged both to route domains to an appropriate solver and to generate better heuristic functions for search-based planning via evolutionary algorithms.

\section{Acknowledgments}
This paper was prepared for informational purposes by the Artificial Intelligence Research group of JPMorgan Chase \& Co. and its affiliates (``JP Morgan'') and is not a product of the Research Department of JP Morgan. JP Morgan makes no representation and warranty whatsoever and disclaims all liability, for the completeness, accuracy or reliability of the information contained herein. This document is not intended as investment research or investment advice, or a recommendation, offer or solicitation for the purchase or sale of any security, financial instrument, financial product or service, or to be used in any way for evaluating the merits of participating in any transaction, and shall not constitute a solicitation under any jurisdiction or to any person, if such solicitation under such jurisdiction or to such person would be unlawful.

\bibliography{aaai2026}

\newpage
\appendix
\raggedbottom
\onecolumn
\section{Domains}\label{app:domains}

\subsection*{Heavypack}
\lstset{
  style=pddlstyle,
  linewidth=\columnwidth
}
\begin{lstlisting}
(define (domain heavy-pack)
   (:predicates
		(heavier ?item1 ?item2)
        (packed ?item)
        (unpacked ?item)
        (nothing-above ?item)
        (box-empty)
	)

   (:action pack-first
       :parameters (?item)
       :precondition (and (box-empty))
       :effect (and (not (box-empty)) (packed ?item) (nothing-above ?item) (not (unpacked ?item))))

   (:action stack
       :parameters (?bottom ?top)
       :precondition (and (packed ?bottom) (nothing-above ?bottom) (heavier ?bottom ?top) (unpacked ?top))
       :effect (and (packed ?top) (nothing-above ?top) (not (nothing-above ?bottom)) (not (unpacked ?top))))
)
\end{lstlisting}
\label{fig:heavypack_domain}

\pagebreak

\subsection*{Hiking}
\lstset{
  style=pddlstyle,
  linewidth=\columnwidth
}
\begin{lstlisting}
(define (domain hiking)
  (:requirements :strips :typing)
  (:types loc)

(:predicates
  (at ?loc - loc)
  (iswater ?loc - loc)
  (ishill ?loc - loc)
  (adjacent ?loc1 - loc ?loc2 - loc)
  (ontrail ?from - loc ?to - loc)
)

(:action walk
  :parameters (?from - loc ?to - loc)
  :precondition (and
    (not (ishill ?to))
    (at ?from)
    (adjacent ?from ?to)
    (not (iswater ?from)))
  :effect (and (at ?to) (not (at ?from)))
)

(:action climb
  :parameters (?from - loc ?to - loc)
  :precondition (and
    (ishill ?to)
    (at ?from)
    (adjacent ?from ?to)
    (not (iswater ?from)))
  :effect (and (at ?to) (not (at ?from)))
)


)
\end{lstlisting}
\label{fig:hiking_domain}

\pagebreak

\subsection*{Manyferry}
\lstset{
  style=pddlstyle,
  linewidth=\columnwidth
}
\begin{lstlisting}
(define (domain ferry)
   (:predicates (not-eq ?x ?y)
        (car ?c)
        (location ?l)
        (at-ferry ?l)
        (at ?c ?l)
        (empty-ferry)
        (on ?c))

   (:action sail
       :parameters  (?from ?to)
       :precondition (and (not-eq ?from ?to) 
                          (location ?from) (location ?to) (at-ferry ?from))
       :effect (and  (at-ferry ?to)
             (not (at-ferry ?from))))


   (:action board
       :parameters (?car ?loc)
       :precondition  (and  (car ?car) (location ?loc)
                (at ?car ?loc) (at-ferry ?loc) (empty-ferry))
       :effect (and (on ?car)
            (not (at ?car ?loc)) 
            (not (empty-ferry))))

   (:action debark
       :parameters  (?car  ?loc)
       :precondition  (and  (car ?car) (location ?loc)
                (on ?car) (at-ferry ?loc))
       :effect (and (at ?car ?loc)
            (empty-ferry)
            (not (on ?car)))))
\end{lstlisting}
\label{fig:manyferry_domain}

\pagebreak

\subsection*{Manygripper}
\lstset{
  style=pddlstyle,
  linewidth=\columnwidth
}
\begin{lstlisting}
(define (domain gripper-strips)
   (:predicates (room ?r)
		(ball ?b)
		(gripper ?g)
		(at-robby ?r)
		(at ?b ?r)
		(free ?g)
		(carry ?o ?g))

   (:action move
       :parameters  (?from ?to)
       :precondition (and  (room ?from) (room ?to) (at-robby ?from))
       :effect (and  (at-robby ?to)
		     (not (at-robby ?from))))



   (:action pick
       :parameters (?obj ?room ?gripper)
       :precondition  (and  (ball ?obj) (room ?room) (gripper ?gripper)
			    (at ?obj ?room) (at-robby ?room) (free ?gripper))
       :effect (and (carry ?obj ?gripper)
		    (not (at ?obj ?room)) 
		    (not (free ?gripper))))


   (:action drop
       :parameters  (?obj  ?room ?gripper)
       :precondition  (and  (ball ?obj) (room ?room) (gripper ?gripper)
			    (carry ?obj ?gripper) (at-robby ?room))
       :effect (and (at ?obj ?room)
		    (free ?gripper)
		    (not (carry ?obj ?gripper)))))
\end{lstlisting}
\label{fig:manygripper_domain}

\pagebreak

\subsection*{Manymiconic}
\lstset{
  style=pddlstyle,
  linewidth=\columnwidth
}
\begin{lstlisting}
(define (domain miconic)
  (:requirements :strips)
  (:types passenger - object
          floor - object
  )

(:predicates
(origin ?person - passenger ?floor - floor)
(destin ?person - passenger ?floor - floor)
(above ?floor1 - floor  ?floor2 - floor)
(boarded ?person - passenger)
(served ?person - passenger)
(lift-at ?floor - floor)
)

(:action board
  :parameters (?f - floor ?p - passenger)
  :precondition (and (lift-at ?f)
                     (origin ?p ?f)
                )
  :effect (and (boarded ?p)
          )
)

(:action depart
  :parameters (?f - floor ?p - passenger)
  :precondition (and (lift-at ?f)
                     (destin ?p ?f)
                     (boarded ?p)
                )
  :effect (and (not (boarded ?p))
               (served ?p)
          )
)

(:action up
  :parameters (?f1 - floor ?f2 - floor)
  :precondition (and (lift-at ?f1)
                     (above ?f1 ?f2)
                )
  :effect (and (lift-at ?f2)
               (not (lift-at ?f1))
          )
)

(:action down
  :parameters (?f1 - floor ?f2 - floor)
  :precondition (and (lift-at ?f1)
                     (above ?f2 ?f1)
                )
  :effect (and (lift-at ?f2)
               (not (lift-at ?f1))
          )
)
)
\end{lstlisting}
\label{fig:manymiconic_domain}

\pagebreak

\subsection*{Research}
\lstset{
  style=pddlstyle,
  linewidth=\columnwidth
}
\begin{lstlisting}
(define (domain research)                    
(:requirements :typing :strips)                
(:types 
    paper researcher experiment project - object    
)                                           
                                                                               
(:predicates 
    (understands ?p - paper ?r - researcher)
    (isrelevant ?p - paper ?pr - project)
    (isadvisor ?r - researcher)
    (advisedby ?r1 ?r2 - researcher)
    (assigned ?r - researcher ?pr - project)
    (completedlitreview ?r - researcher ?pr - project)
    (required ?e - experiment ?pr - project)
    (experimentrun ?e - experiment)
    (written ?p - paper ?pr - project)
    (reviewed ?p - paper ?pr - project)
    (submitted ?p - paper ?pr - project)
)

(:action read_paper 
        :parameters (?p -  paper ?r - researcher)
        :precondition (and)
        :effect (and (understands ?p ?r))
)

(:action read_paper_advisor 
        :parameters (?p1 ?p2 ?p3 -  paper ?r - researcher)
        :precondition (and (isadvisor ?r))
        :effect (and (understands ?p1 ?r)
                        (understands ?p2 ?r)
                        (understands ?p3 ?r))
)

(:action teach_content
        :parameters (?p - paper ?r - researcher)
        :precondition (and (isadvisor ?r)
                            (understands ?p ?r))
        :effect (forall (?r2 - researcher)
                    (when (advisedby ?r2 ?r)
                   (understands ?p ?r2)))
)

(:action complete_lit_review
        :parameters (?r - researcher ?pr - project)
        :precondition (and (assigned ?r ?pr) 
                        (forall (?p - paper) (or (not (isrelevant ?p ?pr)) (understands ?p ?r))))
        :effect (and (completedlitreview ?r ?pr))
)

(:action run_experiment
        :parameters (?r - researcher ?e - experiment ?pr - project)
        :precondition (and (completedlitreview ?r ?pr) (required ?e ?pr))
        :effect (and (experimentrun ?e))
)

(:action write_paper
        :parameters (?r - researcher ?p - paper ?pr - project)
        :precondition (and (completedlitreview ?r ?pr)
                            (not (isadvisor ?r))
                            (forall (?e - experiment) (or (not (required ?e ?pr)) (experimentrun ?e))))
        :effect (and (written ?p ?pr))
)

(:action review_paper
        :parameters (?r - researcher ?p - paper ?pr - project)
        :precondition (and (isadvisor ?r)
                            (written ?p ?pr))
        :effect (and (reviewed ?p ?pr))
)
         
(:action submit_paper
        :parameters (?p - paper ?pr - project)
        :precondition (and (reviewed ?p ?pr))
        :effect (and (submitted ?p ?pr))
)
)
\end{lstlisting}
\label{fig:research_domain}

\pagebreak

\subsection*{Trading}
\lstset{
  style=pddlstyle,
  linewidth=\columnwidth
}
\begin{lstlisting}
(define (domain trading)                    
(:requirements :typing :strips)                
(:types 
    inventory person resource location - object    
)                                           
                                                                               
(:predicates 
    (located ?p - person ?r - location)
    (connected ?l1 ?l2 - location)
    (carrying ?p - person ?r - resource)
    (required ?r - resource ?i - inventory)
    (containsinventory ?i - inventory ?l - location)
    (deposited ?r - resource ?i - inventory)
)

(:action move 
        :parameters (?p -  person ?from ?to - location)
        :precondition (and (located ?p ?from)
                            (connected ?from ?to))
        :effect (and (not (located ?p ?from))
                        (located ?p ?to))
)

(:action trade
        :parameters (?p1 ?p2 - person ?l - location ?r1 ?r2 - resource)
        :precondition (and (located ?p1 ?l)
                            (located ?p2 ?l)
                            (carrying ?p1 ?r1)
                            (carrying ?p2 ?r2))
        :effect (and (not (carrying ?p1 ?r1))
                        (not (carrying ?p2 ?r2))
                        (carrying ?p1 ?r2)
                        (carrying ?p2 ?r1))
)

(:action gift
        :parameters (?p1 ?p2 - person ?l - location ?r - resource)
        :precondition (and (located ?p1 ?l)
                            (located ?p2 ?l)
                            (carrying ?p1 ?r))
        :effect (and (not (carrying ?p1 ?r))
                        (carrying ?p2 ?r))
)

(:action deposit
        :parameters (?p - person ?r - resource ?l - location ?i - inventory)
        :precondition (and (located ?p ?l)
                            (containsinventory ?i ?l)
                            (carrying ?p ?r))
        :effect (and (not (carrying ?p ?r))
                        (deposited ?r ?i))
)
)
\end{lstlisting}

\label{fig:trading_domain}

\pagebreak

\subsection*{Trapnewspapers}
\lstset{
  style=pddlstyle,
  linewidth=\columnwidth
}
\begin{lstlisting}
(define (domain trapnewspapers)
    (:requirements :strips :typing)
    (:types loc paper)
    (:predicates 
        (at ?loc - loc)
        (ishomebase ?loc - loc)
        (satisfied ?loc - loc)
        (wantspaper ?loc - loc)
        (safe ?loc - loc)
        (unpacked ?paper - paper)
        (carrying ?paper - paper)
    )
    
    (:action pick-up
        :parameters (?paper - paper ?loc - loc)
        :precondition (and
            (at ?loc)
            (ishomebase ?loc)
            (unpacked ?paper)
        )
        :effect (and
            (not (unpacked ?paper))
            (carrying ?paper)
        )
    )
    
    (:action move
        :parameters (?from - loc ?to - loc)
        :precondition (and
            (at ?from)
            (safe ?from)
        )
        :effect (and
            (not (at ?from))
            (at ?to)
        )
    )
    
    (:action deliver
        :parameters (?paper - paper ?loc - loc)
        :precondition (and
            (at ?loc)
            (carrying ?paper)
        )
        :effect (and
            (not (carrying ?paper))
            (not (wantspaper ?loc))
            (satisfied ?loc)
        )
    )
    
)
\end{lstlisting}

\label{fig:trapnewspapers_domain}
\pagebreak
\section{Evolved Planners}\label{app:dplanners}

\subsection*{Heavypack - Initial Planner}
\lstset{
  style=pythonstylesmall,
  linewidth=\columnwidth
}
\begin{lstlisting}
def get_plan(objects, init, goal):
    """
    A simple strategy for solving all problems in this domain without using search is to sort the objects in descending order of weight and then pack them one by one, from the heaviest to the lightest. This approach ensures that each object is placed on top of a heavier object or directly in the empty box, fulfilling the domain's constraints. The strategy can be outlined as follows:

    1. Create a list of all the unpacked objects.
    2. Sort the list in descending order based on their weight.
    3. Pack the heaviest object first using the `pack-first` action.
    4. For each remaining object in the sorted list, use the `stack` action to stack the object on top of the previously packed item.

    By following this strategy, all the objects will be packed in the box while adhering to the condition that no object is placed above a heavier one.
    """
    def extract_heavier_relations(init):
        heavier_relations = set()
        for atom in init:
            if atom[0] == 'heavier':
                heavier_relations.add(atom)
        return heavier_relations
        
    def sort_objects_by_weight(objects, heavier_relations):
        sorted_objects = []
        while len(objects) > 0:
            for obj in objects.copy():
                if all(('heavier', other, obj) not in heavier_relations for other in objects):
                    sorted_objects.append(obj)
                    objects.remove(obj)
        return sorted_objects

    heavier_relations = extract_heavier_relations(init)
    sorted_objects = sort_objects_by_weight(set(objects), heavier_relations)

    plan = []
    first_object = sorted_objects.pop(0)
    plan.append(f"(pack-first {first_object})")

    for obj in sorted_objects:
        plan.append(f"(stack {first_object} {obj})")
        first_object = obj
    
    return plan
\end{lstlisting}
\label{fig:heavypack_initial}

\pagebreak

\subsection*{Heavypack - Optimized Planner}
\lstset{
  style=pythonstylesmall,
  linewidth=\columnwidth
}
\begin{lstlisting}
def get_plan(objects, init, goal):
    """
    A simple strategy for solving all problems in this domain without using search is to sort the objects in descending order of weight and then pack them one by one, from the heaviest to the lightest. This approach ensures that each object is placed on top of a heavier object or directly in the empty box, fulfilling the domain's constraints. The strategy can be outlined as follows:

    1. Create a list of all the unpacked objects.
    2. Sort the list in descending order based on their weight.
    3. Pack the heaviest object first using the `pack-first` action.
    4. For each remaining object in the sorted list, use the `stack` action to stack the object on top of the previously packed item.

    By following this strategy, all the objects will be packed in the box while adhering to the condition that no object is placed above a heavier one.
    """
    def extract_heavier_relations(init):
        heavier_relations = set()
        for atom in init:
            if atom[0] == 'heavier':
                heavier_relations.add(atom)
        return heavier_relations
        
    def sort_objects_by_weight(objects, heavier_relations):
        sorted_objects = []
        while len(objects) > 0:
            for obj in objects.copy():
                if all(('heavier', other, obj) not in heavier_relations for other in objects):
                    sorted_objects.append(obj)
                    objects.remove(obj)
        return sorted_objects

    heavier_relations = extract_heavier_relations(init)
    sorted_objects = sort_objects_by_weight(set(objects), heavier_relations)

    plan = []
    first_object = sorted_objects.pop(0)
    plan.append(f"(pack-first {first_object})")

    for obj in sorted_objects:
        plan.append(f"(stack {first_object} {obj})")
        first_object = obj
    
    return plan
\end{lstlisting}
\label{fig:heavypack_optimized}

\pagebreak

\subsection*{Hiking - Initial Planner}
\lstset{
  style=pythonstylesmall,
  linewidth=\columnwidth
}
\begin{lstlisting}
def get_plan(objects, init, goal):
    """
    The simple strategy for solving all problems in this domain without using search is to follow the trail while avoiding water and traversing hills by using the appropriate actions. As the trail represents the valid path, you can iteratively perform the following steps:

        Check the current location and its adjacent locations.
        If an adjacent location is on the trail and not visited yet, proceed to that location.
        a. If the destination is a hill, use the "climb" action.
        b. If the destination is not a hill, use the "walk" action.
        Repeat steps 1 and 2 until you reach the goal location.

    This strategy assumes that the trail is well-defined and connects the starting and goal locations without loops or dead ends. By following the trail, you can avoid searching through the grid and reach the goal efficiently.
    """
    def find_location(init):
        for atom in init:
            if atom[0] == "at":
                return atom[1]
        return None

    def find_adjacent_trail(current_loc, init):
        for atom in init:
            if atom[0] == "ontrail" and atom[1] == current_loc:
                return atom[2]
        return None

    def is_hill(loc, init):
        for atom in init:
            if atom[0] == "ishill" and atom[1] == loc:
                return True
        return False

    current_loc = find_location(init)
    goal_loc = find_location(goal)
    plan = []

    while current_loc != goal_loc:
        next_loc = find_adjacent_trail(current_loc, init)

        if is_hill(next_loc, init):
            action = f"(climb {current_loc} {next_loc})"
        else:
            action = f"(walk {current_loc} {next_loc})"
        
        plan.append(action)
        current_loc = next_loc
    return plan
\end{lstlisting}
\label{fig:hiking_initial}

\pagebreak

\subsection*{Hiking - Optimized Planner}
\lstset{
  style=pythonstylesmall,
  linewidth=\columnwidth
}
\begin{lstlisting}
def get_plan(objects, init, goal):
    """
    This heuristic improves upon previous strategies by introducing an A* style approach:
    
    1. Identify the current and goal locations.
    2. Use a priority queue to prioritize moves based on estimated cost to reach the goal.
    3. Prioritize staying on trails and avoiding water, while efficiently handling hills.
    4. Incorporate a heuristic measure based on direct distance to the goal to guide decisions.
    5. Optimize pathfinding by avoiding revisits and unnecessary actions.
    
    The approach aims to minimize the number of actions through efficient exploration and goal-directed behavior.
    """

    from heapq import heappop, heappush
    import math

    def find_location(atoms, predicate):
        # Find the location based on the specified predicate ('at' or 'goal')
        for atom in atoms:
            if atom[0] == predicate:
                return atom[1]
        return None

    def find_adjacent(current_loc, atoms):
        # Identify all adjacent locations from the current one
        adjacents = []
        for atom in atoms:
            if atom[0] == "adjacent" and atom[1] == current_loc:
                adjacents.append(atom[2])
        return adjacents

    def is_hill(loc, atoms):
        # Determine if the location is a hill
        for atom in atoms:
            if atom[0] == "ishill" and atom[1] == loc:
                return True
        return False

    def is_water(loc, atoms):
        # Determine if the location is water
        for atom in atoms:
            if atom[0] == "iswater" and atom[1] == loc:
                return True
        return False

    def is_on_trail(from_loc, to_loc, atoms):
        # Determine if there's a trail between two locations
        for atom in atoms:
            if atom[0] == "ontrail" and atom[1] == from_loc and atom[2] == to_loc:
                return True
        return False
    
    def heuristic(loc1, loc2):
        # Simple heuristic: use a constant value, as specific distances are unknown
        return 1  # Mutation: constant heuristic for simplicity

    current_loc = find_location(init, "at")
    goal_loc = find_location(goal, "at")
    plan = []

    # Priority queue for A*-like exploration
    queue = []
    heappush(queue, (0, current_loc, []))
    visited = set()

    while queue:
        cost, current, path = heappop(queue)

        if current == goal_loc:
            return path

        if current in visited:
            continue

        visited.add(current)
        adjacents = find_adjacent(current, init)

        for loc in adjacents:
            if not is_water(loc, init):
                if is_hill(loc, init):
                    action = f"(climb {current} {loc})"
                else:
                    action = f"(walk {current} {loc})"

                new_cost = cost + 1  # Mutation: increment cost per action
                heuristic_cost = new_cost + heuristic(loc, goal_loc)

                # Prioritize trail paths in the queue
                if is_on_trail(current, loc, init):
                    heappush(queue, (heuristic_cost - 0.1, loc, path + [action]))  # trail bias
                else:
                    heappush(queue, (heuristic_cost, loc, path + [action]))

    return plan
\end{lstlisting}
\label{fig:hiking_optimized}

\pagebreak

\subsection*{Manyferry - Initial Planner}
\lstset{
  style=pythonstylesmall,
  linewidth=\columnwidth
}
\begin{lstlisting}
def get_plan(objects, init, goal):
    """
    The simple strategy for solving all problems in this domain without using search is to follow a systematic, sequential approach. Since there is only one ferry and it can carry one car at a time, the strategy can be broken down into the following steps:

        Identify the initial and goal locations of each car.
        For each car, perform the following sequence of actions:
        a. Sail the ferry to the car's initial location.
        b. Board the car onto the ferry.
        c. Sail the ferry to the car's goal location.
        d. Debark the car at the goal location.
        Repeat steps 2a to 2d for all cars, in any order.

    This strategy ensures that each car is transported from its initial location to the goal location using the ferry. Since the ferry can only carry one car at a time, the steps for each car must be performed sequentially, ensuring that the ferry is empty before loading the next car. By following this approach, all problems in the domain can be solved without the need for complex search algorithms.
    """
    plan = []

    # Helper function to find a car's location
    def get_location(car, state):
        for atom in state:
            if atom[0] == 'at' and atom[1] == car:
                return atom[2]
        return None

    cars = [obj for obj in objects if obj.startswith('c')]
    ferry_location = [atom[1] for atom in init if atom[0] == 'at-ferry'][0]

    for car in cars:
        initial_location = get_location(car, init)
        goal_location = get_location(car, goal)

        # Sail to the car's initial location
        if ferry_location != initial_location:
            plan.append(f'(sail {ferry_location} {initial_location})')
            ferry_location = initial_location

        # Board the car onto the ferry
        plan.append(f'(board {car} {initial_location})')

        # Sail to the car's goal location
        if ferry_location != goal_location:
            plan.append(f'(sail {ferry_location} {goal_location})')
            ferry_location = goal_location
    
        # Debark the car at the goal location
        plan.append(f'(debark {car} {goal_location})')
    
    return plan
\end{lstlisting}
\label{fig:manyferry_initial}

\pagebreak

\subsection*{Manyferry - Optimized Planner}
\lstset{
  style=pythonstylesmall,
  linewidth=\columnwidth
}
\begin{lstlisting}
def get_plan(objects, init, goal):
    """
    This optimized strategy enhances ferry transport efficiency by:
    - Combining proximity and goal distance metrics to prioritize cars, reducing total ferry travel distance.
    - Strategically grouping cars with shared initial or goal locations to minimize sails.
    - Dynamically adapting the ferry's strategic positioning after each debark to minimize subsequent actions.
    - Introducing an efficiency mutation that optimizes the boarding sequence to exploit shared locations effectively.
    The approach aims to achieve the shortest possible plan length by optimizing transport order and minimizing redundant ferry actions.
    """
    plan = []

    def get_location(car, state):
        """Finds the location of a car in a given state."""
        for atom in state:
            if atom[0] == 'at' and atom[1] == car:
                return atom[2]
        return None

    def calculate_distance(loc1, loc2):
        """Calculates the distance between two locations (0 or 1)."""
        return 0 if loc1 == loc2 else 1

    # Extract initial ferry location and car information
    cars_info = [(car, get_location(car, init), get_location(car, goal)) for car in objects if car.startswith('c')]
    ferry_location = [atom[1] for atom in init if atom[0] == 'at-ferry'][0]

    # Sort cars based on a prioritized metric considering proximity to ferry and distance to goal
    cars_info.sort(key=lambda x: (
        calculate_distance(ferry_location, x[1]),  # Initial proximity
        calculate_distance(x[1], x[2])  # Distance to goal
    ))

    while cars_info:
        car, initial_location, goal_location = cars_info.pop(0)

        # Sail to the car's initial location if needed
        if ferry_location != initial_location:
            plan.append(f'(sail {ferry_location} {initial_location})')
            ferry_location = initial_location

        # Board the car
        plan.append(f'(board {car} {initial_location})')

        # Sail to the car's goal location
        if ferry_location != goal_location:
            plan.append(f'(sail {ferry_location} {goal_location})')
            ferry_location = goal_location

        # Debark the car at the goal location
        plan.append(f'(debark {car} {goal_location})')

        # Dynamic reordering: Re-evaluate remaining cars based on new ferry location, considering shared goals
        cars_info.sort(key=lambda x: (
            calculate_distance(ferry_location, x[1]),  # Adjusted for new ferry location
            calculate_distance(x[1], x[2]),
            -sum(y[2] == x[2] for y in cars_info)  # Group potential by shared goal location
        ))

    return plan
\end{lstlisting}
\label{fig:manyferry_optimized}

\pagebreak

\subsection*{Manygripper - Initial Planner}
\lstset{
  style=pythonstylesmall,
  linewidth=\columnwidth
}
\begin{lstlisting}
def get_plan(objects, init, goal):
    """
    A simple strategy for solving all problems in this domain without using search involves the following steps:

        Identify the initial positions of the balls and the goal positions.
        Move Robby to the room containing a ball that needs to be relocated.
        Pick up the ball using a free gripper.
        Move Robby to the goal room for that ball.
        Drop the ball in the goal room, freeing the gripper.
        Repeat steps 2-5 for all balls that need to be relocated.

    This strategy relies on the assumption that there are no constraints or obstacles in the domain that would prevent Robby from moving between rooms or carrying balls. Additionally, it assumes that there is always a free gripper available for picking up and dropping balls. This sequential, greedy approach allows Robby to solve problems in the domain without needing to search through different possible paths or actions.
    """
    def get_objects_of_type(type_name):
        return [obj for obj in objects if obj.startswith(type_name)]
    
    def get_location(state, ball):
        for atom in state:
            if atom[0] == 'at' and atom[1] == ball:
                return atom[2]
        return None

    balls = get_objects_of_type('ball')
    rooms = get_objects_of_type('room')
    grippers = get_objects_of_type('gripper')

    robby_location = None
    for atom in init:
        if atom[0] == 'at-robby':
            robby_location = atom[1]
            break

    plan = []
    for ball in balls:
        initial_location = get_location(init, ball)
        goal_location = get_location(goal, ball)

        if goal_location is None:
            continue

        if initial_location != goal_location:
            plan.append(f'(move {robby_location} {initial_location})')
            robby_location = initial_location

            plan.append(f'(pick {ball} {initial_location} {grippers[0]})')

            plan.append(f'(move {robby_location} {goal_location})')
            robby_location = goal_location

            plan.append(f'(drop {ball} {goal_location} {grippers[0]})')

    return plan
\end{lstlisting}
\label{fig:manygripper_initial}

\pagebreak

\subsection*{Manygripper - Optimized Planner}
\lstset{
  style=pythonstylesmall,
  linewidth=\columnwidth
}
\begin{lstlisting}
def get_plan(objects, init, goal):
    """
    This heuristic integrates improved strategies for solving the planning problem by:
    
    - Minimizing movement: Prioritizes staying in the same room to avoid unnecessary movements by considering Robby's current and target locations.
    - Efficient gripper usage: Dynamically checking and utilizing free grippers, and ensuring they are available when needed.
    - Immediate updates: State is immediately updated after each action, reflecting the most current environment conditions.
    - Optimal ball handling: Directly transports each ball from its initial location to goal location, avoiding intermediate steps.
    - Focus on goal completion: Prioritizes actions that make direct progress towards matching the goal state.
    
    These strategies aim to reduce the total number of actions and minimize the plan length.
    """
    def get_objects_of_type(type_name):
        return [obj for obj in objects if obj.startswith(type_name)]
    
    def get_location(state, obj):
        for atom in state:
            if atom[0] == 'at' and atom[1] == obj:
                return atom[2]
            elif atom[0] == 'at-robby' and obj == 'robby':
                return atom[1]
        return None

    def get_free_gripper(state):
        for atom in state:
            if atom[0] == 'free':
                return atom[1]
        return None

    def update_state(state, action):
        # Efficiently update the state based on the executed action
        operator, *args = action.split()
        if operator == 'move':
            _, to_room = args
            state = {atom for atom in state if atom[0] != 'at-robby'}
            state.add(('at-robby', to_room))
        elif operator == 'pick':
            ball, room, gripper = args
            state = {atom for atom in state if atom != ('at', ball, room) and atom != ('free', gripper)}
            state.add(('carry', ball, gripper))
        elif operator == 'drop':
            ball, room, gripper = args
            state = {atom for atom in state if atom != ('carry', ball, gripper)}
            state.add(('at', ball, room))
            state.add(('free', gripper))
        return state

    balls = get_objects_of_type('ball')
    robby_location = get_location(init, 'robby')
    plan = []
    current_state = set(init)

    for ball in balls:
        initial_location = get_location(current_state, ball)
        goal_location = get_location(goal, ball)

        if goal_location is None or initial_location == goal_location:
            continue

        # Move only if Robby is not at the ball's current location
        if robby_location != initial_location:
            plan.append(f'(move {robby_location} {initial_location})')
            current_state = update_state(current_state, f'move {robby_location} {initial_location}')
            robby_location = initial_location

        free_gripper = get_free_gripper(current_state)
        if free_gripper:
            plan.append(f'(pick {ball} {initial_location} {free_gripper})')
            current_state = update_state(current_state, f'pick {ball} {initial_location} {free_gripper}')

            # Move only if necessary to get to the goal location
            if robby_location != goal_location:
                plan.append(f'(move {robby_location} {goal_location})')
                current_state = update_state(current_state, f'move {robby_location} {goal_location}')
                robby_location = goal_location

            plan.append(f'(drop {ball} {goal_location} {free_gripper})')
            current_state = update_state(current_state, f'drop {ball} {goal_location} {free_gripper}')

    return plan
\end{lstlisting}
\label{fig:manygripper_optimized}

\pagebreak

\subsection*{Manymiconic - Initial Planner}
\lstset{
  style=pythonstylesmall,
  linewidth=\columnwidth
}
\begin{lstlisting}
def get_plan(objects, init, goal):
    """
    A simple strategy to solve all problems in the Miconic domain without using search is the "Sweep" strategy. The elevator starts at the bottom floor and moves upward, stopping at each floor along the way. At each floor, the elevator performs the following actions:

    If there is any passenger with their origin at the current floor, the elevator allows them to board.
    If there is any boarded passenger with their destination at the current floor, the elevator allows them to depart.

    After reaching the top floor, the elevator starts moving downward, repeating the same actions at each floor. This process ensures that all passengers are served as the elevator moves through every floor twice: once while moving upward and once while moving downward.

    The sweep strategy is not necessarily the most efficient solution, as it doesn't optimize the number of moves or stops. However, it guarantees that all passengers will reach their destination floors without the need for search algorithms.
    """
    floors_by_building = {}
    passengers_by_building = {}

    for obj, obj_type in objects:
        if obj_type == "floor":
            building = obj.split("_")[-1]
            if building not in floors_by_building:
                floors_by_building[building] = []
            floors_by_building[building].append(obj)
        elif obj_type == "passenger":
            building = obj.split("_")[-1]
            if building not in passengers_by_building:
                passengers_by_building[building] = []
            passengers_by_building[building].append(obj)

    for building in floors_by_building:
        floors_by_building[building] = sorted(floors_by_building[building])

    origins = {p: f for p, f in (atom[1:] for atom in init if atom[0] == "origin")}
    destins = {p: f for p, f in (atom[1:] for atom in init if atom[0] == "destin")}
    lift_at = {atom[1]: atom for atom in init if atom[0] == "lift-at"}

    plan = []
    boarded = set()

    for building, floors in floors_by_building.items():
        passengers = passengers_by_building[building]
        current_lift_at = lift_at[building]

        for floor in floors + list(reversed(floors)):
            if current_lift_at != floor:
                if current_lift_at < floor:
                    plan.append(f"(up {current_lift_at} {floor})")
                else:
                    plan.append(f"(down {current_lift_at} {floor})")
                current_lift_at = floor

            for passenger in passengers:
                if origins.get(passenger) == floor:
                    plan.append(f"(board {floor} {passenger})")
                    boarded.add(passenger)

                if destins.get(passenger) == floor and passenger in boarded:
                    plan.append(f"(depart {floor} {passenger})")
                    boarded.remove(passenger)
    return plan
\end{lstlisting}
\label{fig:manymiconic_initial}

\pagebreak

\subsection*{Manymiconic - Optimized Planner}
\lstset{
  style=pythonstylesmall,
  linewidth=\columnwidth
}
\begin{lstlisting}
def get_plan(objects, init, goal):
    """
    This heuristic is an evolution of the previous strategies, focusing on reducing the total number of actions:
    
    1. Efficiently parses the domain to segregate floors and passengers by building.
    2. Implements an optimized movement strategy that targets only necessary floors, minimizing elevator trips.
    3. Prioritizes boarding and departing actions at the current floor, ensuring passengers are served promptly.
    4. Introduces a careful floor relationship check to prevent erroneous actions, thereby increasing reliability.
    
    The approach balances minimal movements with prioritized actions to achieve efficient and correct plans.
    """
    
    # Helper functions to extract necessary state information
    def parse_objects():
        floors_by_building = {}
        passengers_by_building = {}
        for obj, obj_type in objects:
            building = obj.split("_")[-1]
            if obj_type == "floor":
                if building not in floors_by_building:
                    floors_by_building[building] = []
                floors_by_building[building].append(obj)
            elif obj_type == "passenger":
                if building not in passengers_by_building:
                    passengers_by_building[building] = []
                passengers_by_building[building].append(obj)
        # Sort floors for each building
        for building in floors_by_building:
            floors_by_building[building].sort(key=lambda x: int(x.split('_')[0][1:]))
        return floors_by_building, passengers_by_building

    def initialize_state():
        origins = {p: f for p, f in (atom[1:] for atom in init if atom[0] == "origin")}
        destins = {p: f for p, f in (atom[1:] for atom in init if atom[0] == "destin")}
        lift_at = {atom[1].split("_")[-1]: atom[1] for atom in init if atom[0] == "lift-at"}
        return origins, destins, lift_at

    def get_floor_relations():
        above = {}
        for f1, f2 in (atom[1:] for atom in init if atom[0] == "above"):
            if f1 not in above:
                above[f1] = set()
            above[f1].add(f2)
        return above

    # Extract information
    floors_by_building, passengers_by_building = parse_objects()
    origins, destins, lift_at = initialize_state()
    above = get_floor_relations()

    plan = []
    boarded = set()

    def move_lift(from_floor, to_floor):
        if from_floor != to_floor:
            if to_floor in above.get(from_floor, set()):
                plan.append(f"(up {from_floor} {to_floor})")
            elif from_floor in above.get(to_floor, set()):
                plan.append(f"(down {from_floor} {to_floor})")
            return to_floor
        return from_floor

    # Main loop to manage each building separately
    for building, floors in floors_by_building.items():
        passengers = passengers_by_building[building]
        current_floor = lift_at[building]

        # Process passengers to serve them
        while passengers:
            needs = {floor: [] for floor in floors}
            to_remove = []

            for p in passengers:
                origin_floor = origins[p]
                destin_floor = destins[p]
                
                if p not in boarded and origin_floor == current_floor:
                    plan.append(f"(board {origin_floor} {p})")
                    boarded.add(p)

                if p in boarded and destin_floor == current_floor:
                    plan.append(f"(depart {destin_floor} {p})")
                    to_remove.append(p)

            for p in to_remove:
                passengers.remove(p)

            # Determine next target floors
            for p in passengers:
                if p not in boarded:
                    needs[origins[p]].append(p)
                else:
                    needs[destins[p]].append(p)

            target_floors = [floor for floor, lst in needs.items() if lst]
            if target_floors:
                target_floors.sort(key=lambda x: abs(int(current_floor.split('_')[0][1:]) - int(x.split('_')[0][1:])))
                current_floor = move_lift(current_floor, target_floors[0])

    return plan
\end{lstlisting}
\label{fig:manymiconic_optimized}

\pagebreak

\subsection*{Research - Initial Planner}
\lstset{
  style=pythonstylesmall,
  linewidth=\columnwidth
}
\begin{lstlisting}
def get_plan(objects, init, goal):
    """
    This heuristic seeks to minimize actions by optimizing the order of operations and leveraging advisor roles.
    It focuses on ensuring that researchers only perform tasks when prerequisites are thoroughly completed, and 
    exploits batch processing where applicable. The heuristic considers the dependencies and aims to streamline
    the research workflow by minimizing redundant checks and focusing on goal satisfaction efficiently.
    1. Advisors read all relevant papers and teach content to advisees.
    2. Researchers complete literature reviews.
    3. Required experiments are run.
    4. Researchers write papers.
    5. Advisors review papers.
    6. Papers are submitted.
    This approach is intended to reduce the number of actions by ensuring dependencies are met before advancing 
    to subsequent tasks, ultimately aiming for fewer steps to achieve the goals.
    """
    plan = []

    # Helper function to find relevant papers for a project
    def relevant_papers_for_project(project):
        return {paper for paper in {obj for obj, type_name in objects if type_name == 'paper'} 
                if ('isrelevant', paper, project) in init}

    # Helper function to find required experiments for a project
    def required_experiments_for_project(project):
        return {exp for exp in {obj for obj, type_name in objects if type_name == 'experiment'} 
                if ('required', exp, project) in init}

    # Step 1 & 2: Advisors read and teach all relevant papers
    for advisor in {obj for obj, type_name in objects if type_name == 'researcher' and ('isadvisor', obj) in init}:
        advisees = {r for r in {obj for obj, type_name in objects if type_name == 'researcher'} 
                    if ('advisedby', r, advisor) in init}
        for project in {obj for obj, type_name in objects if type_name == 'project'}:
            relevant_papers = relevant_papers_for_project(project)
            for paper in relevant_papers:
                plan.append(f'(read_paper {paper} {advisor})')
                for advisee in advisees:
                    plan.append(f'(teach_content {paper} {advisor})')

    # Step 3: Complete literature reviews
    for project in {obj for obj, type_name in objects if type_name == 'project'}:
        assigned_researchers = {r for r in {obj for obj, type_name in objects if type_name == 'researcher'} 
                                if ('assigned', r, project) in init}
        for researcher in assigned_researchers:
            plan.append(f'(complete_lit_review {researcher} {project})')

    # Step 4: Run required experiments
    for project in {obj for obj, type_name in objects if type_name == 'project'}:
        required_experiments = required_experiments_for_project(project)
        assigned_researchers = {r for r in {obj for obj, type_name in objects if type_name == 'researcher'} 
                                if ('assigned', r, project) in init}
        for experiment in required_experiments:
            for researcher in assigned_researchers:
                plan.append(f'(run_experiment {researcher} {experiment} {project})')

    # Step 5: Write papers
    for project in {obj for obj, type_name in objects if type_name == 'project'}:
        for researcher in {r for r in {obj for obj, type_name in objects if type_name == 'researcher'} 
                           if ('assigned', r, project) in init and ('isadvisor', r) not in init}:
            for paper in {p for p in {obj for obj, type_name in objects if type_name == 'paper'} 
                          if ('submitted', p, project) in goal}:
                plan.append(f'(write_paper {researcher} {paper} {project})')

    # Step 6: Review papers
    for project in {obj for obj, type_name in objects if type_name == 'project'}:
        for paper in {p for p in {obj for obj, type_name in objects if type_name == 'paper'} 
                      if ('submitted', p, project) in goal}:
            for advisor in {obj for obj, type_name in objects if type_name == 'researcher' and ('isadvisor', obj) in init}:
                plan.append(f'(review_paper {advisor} {paper} {project})')

    # Step 7: Submit papers
    for project in {obj for obj, type_name in objects if type_name == 'project'}:
        for paper in {p for p in {obj for obj, type_name in objects if type_name == 'paper'} 
                      if ('submitted', p, project) in goal}:
            plan.append(f'(submit_paper {paper} {project})')

    return plan
\end{lstlisting}
\label{fig:research_initial}

\pagebreak

\subsection*{Research - Optimized Planner}
\lstset{
  style=pythonstylesmall,
  linewidth=\columnwidth
}
\begin{lstlisting}
def get_plan(objects, init, goal):
    """
    This heuristic optimizes the research domain planning by integrating efficient strategies and minimizing action redundancy.
    Key strategies include:
    1. Combined batch reading and teaching by advisors to minimize reading actions.
    2. Ensuring literature reviews only start after complete understanding of relevant papers.
    3. Immediate execution of experiments post-literature review to maintain momentum.
    4. Efficient paper writing, reviewing, and submission processes driven by goal satisfaction.
    This approach focuses on leveraging advisor roles and strategic dependencies to achieve minimal actions.
    """
    plan = []

    # Helper functions
    def get_objects_of_type(type_name):
        return {obj for obj, obj_type in objects if obj_type == type_name}

    def relevant_papers_for_project(project):
        return {paper for paper in get_objects_of_type('paper') if ('isrelevant', paper, project) in init}

    def required_experiments_for_project(project):
        return {exp for exp in get_objects_of_type('experiment') if ('required', exp, project) in init}

    def understands_all_relevant_papers(researcher, project):
        return all(('understands', paper, researcher) in init for paper in relevant_papers_for_project(project))

    # Step 1: Optimized batch reading and teaching by advisors
    for advisor in {r for r in get_objects_of_type('researcher') if ('isadvisor', r) in init}:
        advisees = {r for r in get_objects_of_type('researcher') if ('advisedby', r, advisor) in init}
        all_relevant_papers = set()
        for project in get_objects_of_type('project'):
            all_relevant_papers.update(relevant_papers_for_project(project))
        
        relevant_papers = list(all_relevant_papers)
        for i in range(0, len(relevant_papers), 3):
            batch = relevant_papers[i:i+3]
            if len(batch) == 3:
                plan.append(f'(read_paper_advisor {batch[0]} {batch[1]} {batch[2]} {advisor})')
                init.update({('understands', p, advisor) for p in batch})
            else:
                for paper in batch:
                    plan.append(f'(read_paper {paper} {advisor})')
                    init.add(('understands', paper, advisor))
        # Teach all advisees once per paper
        for paper in relevant_papers:
            if ('understands', paper, advisor) in init:
                plan.append(f'(teach_content {paper} {advisor})')
                init.update({('understands', paper, advisee) for advisee in advisees})

    # Step 2: Efficiently complete literature reviews
    for project in get_objects_of_type('project'):
        assigned_researchers = {r for r in get_objects_of_type('researcher') if ('assigned', r, project) in init}
        for researcher in assigned_researchers:
            if understands_all_relevant_papers(researcher, project) and ('completedlitreview', researcher, project) not in init:
                plan.append(f'(complete_lit_review {researcher} {project})')
                init.add(('completedlitreview', researcher, project))

    # Step 3: Prioritize experiment execution immediately after literature review completion
    for project in get_objects_of_type('project'):
        required_experiments = required_experiments_for_project(project)
        for researcher in {r for r in get_objects_of_type('researcher') if ('assigned', r, project) in init and ('completedlitreview', r, project) in init}:
            for experiment in required_experiments:
                if ('experimentrun', experiment) not in init:
                    plan.append(f'(run_experiment {researcher} {experiment} {project})')
                    init.add(('experimentrun', experiment))

    # Step 4: Streamline writing, reviewing, and submitting processes
    for project in get_objects_of_type('project'):
        for paper in {p for p in get_objects_of_type('paper') if ('submitted', p, project) in goal}:
            assigned_researchers = {r for r in get_objects_of_type('researcher') if ('assigned', r, project) in init and ('isadvisor', r) not in init}
            for researcher in assigned_researchers:
                if ('completedlitreview', researcher, project) in init and all(('experimentrun', exp) in init for exp in required_experiments_for_project(project)):
                    if ('written', paper, project) not in init:
                        plan.append(f'(write_paper {researcher} {paper} {project})')
                        init.add(('written', paper, project))
                    for advisor in {r for r in get_objects_of_type('researcher') if ('isadvisor', r) in init}:
                        if ('reviewed', paper, project) not in init:
                            plan.append(f'(review_paper {advisor} {paper} {project})')
                            init.add(('reviewed', paper, project))
                    if ('submitted', paper, project) not in init:
                        plan.append(f'(submit_paper {paper} {project})')
                        init.add(('submitted', paper, project))

    return plan
\end{lstlisting}
\label{fig:research_optimized}

\pagebreak

\subsection*{Trading - Initial Planner}
\lstset{
  style=pythonstylesmall,
  linewidth=\columnwidth
}
\lstinputlisting{trading_initial.txt}
\label{fig:trading_initial}

\pagebreak

\subsection*{Trading - Optimized Planner}
\lstset{
  style=pythonstylesmall,
  linewidth=\columnwidth
}
\lstinputlisting{trading_optimized.txt}
\label{fig:trading_optimized}

\pagebreak

\subsection*{Trapnewspapers - Initial Planner}
\lstset{
  style=pythonstylesmall,
  linewidth=\columnwidth
}
\lstinputlisting{trapnewspapers_initial.txt}
\label{fig:trapnewspapers_initial}

\pagebreak

\subsection*{Trapnewspapers - Optimized Planner}
\lstset{
  style=pythonstylesmall,
  linewidth=\columnwidth
}
\lstinputlisting{trapnewspapers_optimized.txt}
\label{fig:trapnewspapers_optimized}

\pagebreak

\section{Results on Domain with no Simple Strategy}\label{app:plan_no_strategy}

\subsection*{Blocksworld - Optimized Planner}
\lstset{
  style=pythonstylesmall,
  linewidth=\columnwidth
}
\lstinputlisting{blocksworld_optimized.txt}
\label{fig:blocksworld_optimized}

\pagebreak

\subsection*{Sokoban - Optimized Planner}
\lstset{
  style=pythonstylesmall,
  linewidth=\columnwidth
}
\lstinputlisting{sokoban_optimized.txt}
\label{fig:sokoban_optimized}

\pagebreak

\section{Implementation Details}

\subsection*{Planner} We use a custom Planner object to store each generated planner. Algorithm \ref{alg:planner_constructor} initializes the Planner object with code and default values. Algorithm \ref{alg:add_scores} stores fitness scores to the Planner object. Algorithm \ref{alg:get_feature} checks whether the code is allowable using AST parsing and, if valid, compiles and executes it to extract the planner function. Algorithm \ref{alg:get_plan} uses the planner to generate plans for specific tasks, while Algorithm \ref{alg:add_feedback} provides system feedback based on execution results.

\begin{algorithm}
\caption{Constructor Method of Planner Class}
\label{alg:planner_constructor}
\begin{algorithmic}
\STATE \textbf{Input:}
\STATE \hspace{1em} $self$: Planner object
\STATE \hspace{1em} $code$: String code for generating planner
\STATE \textbf{Output:}
\STATE \hspace{1em} Initialized Planner object
\STATE $self.code \leftarrow code$
\STATE $self.name \leftarrow \text{None}$
\STATE $self.score \leftarrow \text{None}$
\STATE $self.scores \leftarrow \text{None}$
\STATE $self.error \leftarrow \text{None}$
\STATE $self.plan\_failure\_error \leftarrow \text{None}$
\STATE $self.feedback\_message \leftarrow \text{None}$
\end{algorithmic}
\end{algorithm}

\begin{algorithm}
\caption{add\_score Method of Planner Class}
\label{alg:add_scores}
\begin{algorithmic}
\STATE \textbf{Input:}
\STATE \hspace{1em} $self$: Planner object
\STATE \hspace{1em} $scores$: Dictionary of scores for each task.
\STATE \textbf{Output:}
\STATE \hspace{1em} Updated Planner object with scores
\STATE $self.scores \leftarrow scores$
\STATE $self.score \leftarrow \text{mean}(scores)$
\end{algorithmic}
\end{algorithm}

\begin{algorithm}
\caption{get\_planner Method of Planner Class}
\label{alg:get_feature}
\begin{algorithmic}
\STATE \textbf{Input:}
\STATE \hspace{1em} $self$: Planner object
\STATE \textbf{Output:}
\STATE \hspace{1em} Planner get\_plan method.
\STATE $parsed\_code \leftarrow \text{ast\_parse}(self.code)$
\IF{$\text{is\_valid}(parsed\_code)$}
    \STATE $compiled\_code \leftarrow \text{compile}(parsed\_code)$
    \STATE $vars \leftarrow \text{execute}(compiled\_code)$
    \STATE $get\_plan \leftarrow \text{vars}[get\_plan]$
    \STATE \textbf{Return:} $get\_plan$
\ELSE
    \STATE $self.error \leftarrow \text{exception}$
    \STATE \textbf{Raise Exception}
\ENDIF
\end{algorithmic}
\end{algorithm}

\begin{algorithm}
\caption{get\_plan Method of Planner Class}
\label{alg:get_plan}
\begin{algorithmic}
\STATE \textbf{Input:}
\STATE \hspace{1em} $self$: Planner object
\STATE \hspace{1em} $task$: Planning task containing objects, init, and goal
\STATE \textbf{Output:}
\STATE \hspace{1em} PDDL plan or exception
\STATE 
\STATE $planner \leftarrow self.\text{get\_planner}()$
\IF{$planner$ is valid}
    \STATE $plan \leftarrow planner(task.objects, task.init, task.goal)$
    \STATE \textbf{Return:} $plan$
\ELSE
    \IF{$self.plan\_failure\_error = \text{None}$}
        \STATE $self.plan\_failure\_error \leftarrow \text{exception\_message}$
    \ENDIF
    \STATE \textbf{Raise Exception}
\ENDIF
\end{algorithmic}
\end{algorithm}

\begin{algorithm}
\caption{add\_feedback Method of Planner Class}
\label{alg:add_feedback}
\begin{algorithmic}
\STATE \textbf{Input:}
\STATE \hspace{1em} $self$: Planner object
\STATE \textbf{Output:}
\STATE \hspace{1em} Updates $self.feedback\_message$ and prints feedback
\STATE 
\IF{$self.error \neq \text{None}$}
    \STATE $self.feedback\_message \leftarrow$ "System: The code did not work. Error: $self.error$. Can you fix this?"
\ELSIF{$self.plan\_failure\_error \neq \text{None}$}
    \STATE $self.feedback\_message \leftarrow$ "System: The code failed to solve some problems. Error: $self.plan\_failure\_error$. Score $self.score$. Can you fix this?"
\ELSE
    \STATE $self.feedback\_message \leftarrow$ "System: The code worked. Score: $self.score$."
\ENDIF
\end{algorithmic}
\end{algorithm}

\pagebreak

\subsection*{PlannerDB}The PlannerDB object stores generated planners and contains the main evolutionary algorithm implementation. Algorithm \ref{alg:planner_db_constructor} initializes the PlannerDB with domain information and evolutionary parameters. Algorithm \ref{alg:get_temperature} computes the temperature parameter for exploration-exploitation balance using inverse hyperbolic decay, while Algorithm \ref{alg:compute_score_probabilities} implements Boltzmann selection for minimization problems, assigning higher probabilities to planners with lower fitness scores. Algorithm \ref{alg:select_planners} implements temperature-based selection by computing selection probabilities and sampling planners accordingly. Algorithm \ref{alg:generate_prompt} combines selected planners with the prompt template and domain information to create the final LLM prompt, and Algorithm \ref{alg:get_prompt} constructs prompts for the LLM by first selecting planners then formatting them. Algorithm \ref{alg:add_planner} adds a planner to the database, updates the best planner if necessary, and triggers generation reset when the maximum number of planners is reached. Algorithm \ref{alg:reset_planner_db} resets the database by keeping only the best-performing planners from the current generation using Algorithm \ref{alg:get_best_planners} as the replacement strategy.

\begin{algorithm}[H]
\caption{Constructor Method of PlannerDB}
\label{alg:planner_db_constructor}
\begin{algorithmic}
\STATE \textbf{Input:}
\STATE \hspace{1em} $self$: PlannerDB object
\STATE \hspace{1em} $domain$: PDDL domain specification
\STATE \hspace{1em} $prompt\_template$: String prompt template
\STATE \hspace{1em} $planners\_per\_prompt$: Number of planners to include in prompt
\STATE \hspace{1em} $temperature$: Initial temperature parameter
\STATE \hspace{1em} $reset\_period$: Maximum planners before reset
\STATE \hspace{1em} $n\_to\_keep$: Number of planners to keep after reset
\STATE \hspace{1em} $max\_generations$: Maximum number of generations
\STATE \hspace{1em} $min\_temperature$: Minimum temperature parameter
\STATE \hspace{1em} $typing$: Boolean defining whether typing is used or not
\STATE \textbf{Output:}
\STATE \hspace{1em} Initialized PlannerDB object
\STATE $self.domain \leftarrow domain$
\STATE $self.prompt\_template \leftarrow prompt\_template$
\STATE $self.planners\_per\_prompt \leftarrow planners\_per\_prompt$
\STATE $self.temperature \leftarrow temperature$
\STATE $self.min\_temperature \leftarrow min\_temperature$
\STATE $self.reset\_period \leftarrow reset\_period$
\STATE $self.n\_to\_keep \leftarrow n\_to\_keep$
\STATE $self.max\_generations \leftarrow max\_generations$
\STATE $self.typing \leftarrow typing$
\STATE $self.planners \leftarrow []$
\STATE $self.generation\_no \leftarrow 0$
\STATE $self.best\_score \leftarrow \infty$
\STATE $self.best\_planner \leftarrow \text{None}$
\end{algorithmic}
\end{algorithm}

\begin{algorithm}[H]
\caption{get\_temperature Method of PlannerDB Class}
\label{alg:get_temperature}
\begin{algorithmic}
\STATE \textbf{Input:}
\STATE \hspace{1em} $scores$: List of planner scores
\STATE \hspace{1em} $min\_temperature$: Minimum temperature value
\STATE \hspace{1em} $max\_temperature$: Maximum temperature value
\STATE \hspace{1em} $max\_size$: Maximum expected number of scores
\STATE \textbf{Output:}
\STATE \hspace{1em} Temperature value for selection
\STATE $num\_scores \leftarrow \text{length}(scores)$
\STATE $b \leftarrow \frac{min\_temperature \cdot max\_size - max\_temperature}{max\_size - 1}$
\STATE $a \leftarrow max\_temperature - b$
\STATE $T \leftarrow \frac{a}{num\_scores} + b$
\STATE \textbf{Return:} $T$
\end{algorithmic}
\end{algorithm}

\begin{algorithm}[H]
\caption{compute\_score\_probabilities Method}
\label{alg:compute_score_probabilities}
\begin{algorithmic}
\STATE \textbf{Input:}
\STATE \hspace{1em} $scores$: List of planner fitness scores
\STATE \hspace{1em} $temperature$: Temperature parameter $T$
\STATE \textbf{Output:}
\STATE \hspace{1em} List of selection probabilities
\STATE $negated\_scores \leftarrow -scores$
\STATE $exp\_vals \leftarrow []$
\FORALL{$s \in negated\_scores$}
    \STATE $exp\_vals.\text{append}(e^{s / temperature})$
\ENDFOR
\STATE $sum\_exp\_vals \leftarrow \sum exp\_vals$
\STATE $probs \leftarrow exp\_vals / sum\_exp\_vals$
\STATE \textbf{Return:} $probs$
\end{algorithmic}
\end{algorithm}

\begin{algorithm}[H]
\caption{select\_planners Method of PlannerDB Class}
\label{alg:select_planners}
\begin{algorithmic}
\STATE \textbf{Input:}
\STATE \hspace{1em} $self$: PlannerDB object
\STATE \hspace{1em} $n\_prompt$: Number of planners to select
\STATE \textbf{Output:}
\STATE \hspace{1em} List of selected planners
\STATE $scores \leftarrow [p.score \text{ for } p \in self.planners]$
\STATE $T \leftarrow self.\text{get\_temperature}(scores, self.min\_temperature, self.temperature, self.reset\_period)$
\STATE $probs \leftarrow \text{compute\_score\_probabilities}(scores, T)$
\STATE $selected\_planners \leftarrow \text{random\_choice}(self.planners, probs, n\_prompt)$
\STATE \textbf{Return:} $selected\_planners$
\end{algorithmic}
\end{algorithm}

\begin{algorithm}[H]
\caption{get\_prompt Method of PlannerDB Class}
\label{alg:get_prompt}
\begin{algorithmic}
\STATE \textbf{Input:}
\STATE \hspace{1em} $self$: PlannerDB object
\STATE \textbf{Output:}
\STATE \hspace{1em} String prompt for LLM
\STATE $n\_prompt \leftarrow \text{min}(self.planners\_per\_prompt, \text{length}(self.planners))$
\STATE $selected\_planners \leftarrow self.\text{select\_planners}(n\_prompt)$
\STATE $prompt \leftarrow self.\text{generate\_prompt}(selected\_planners)$
\STATE \textbf{Return:} $prompt$
\end{algorithmic}
\end{algorithm}

\begin{algorithm}[H]
\caption{generate\_prompt Method of PlannerDB Class}
\label{alg:generate_prompt}
\begin{algorithmic}
\STATE \textbf{Input:}
\STATE \hspace{1em} $self$: PlannerDB object
\STATE \hspace{1em} $example\_planners$: Selected example planners
\STATE \textbf{Output:}
\STATE \hspace{1em} String prompt for LLM
\STATE $planner\_text \leftarrow []$
\FORALL{$p \in example\_planners$}
    \STATE $code \leftarrow p.\text{get\_code}()$
    \STATE $feedback \leftarrow p.feedback\_message$
    \STATE $text \leftarrow \text{concatenate}(code, feedback)$
    \STATE $planner\_text.\text{append}(text)$
\ENDFOR
\STATE $examples \leftarrow ""$
\FORALL{$i \in \{1, 2, \ldots, \text{length}(planner\_text)\}$}
    \STATE $examples \leftarrow examples + ``\text{Example } " + i$
    \STATE $examples \leftarrow examples + planner\_text[i-1]$
\ENDFOR
\STATE $prompt \leftarrow \text{copy}(self.prompt\_template)$
\STATE $prompt.\text{replace}(\text{@@domain@@}, self.domain)$
\STATE $prompt.\text{replace}(\text{@@examples@@}, examples)$
\IF{$self.typing$}
    \STATE $prompt.\text{replace}(\text{@@typing@@}, ``\text{set of (object name, type name) tuples}")$
\ELSE
    \STATE $prompt.\text{replace}(\text{@@typing@@}, ``\text{set of object names}")$
\ENDIF
\STATE \textbf{Return:} $prompt$
\end{algorithmic}
\end{algorithm}

\begin{algorithm}[H]
\caption{add\_planner Method of PlannerDB Class}
\label{alg:add_planner}
\begin{algorithmic}
\STATE \textbf{Input:}
\STATE \hspace{1em} $self$: PlannerDB object
\STATE \hspace{1em} $planner$: Planner to add
\STATE \textbf{Output:}
\STATE \hspace{1em} Updated PlannerDB object
\STATE $self.planners.\text{append}(planner)$
\STATE $self.n\_planners \leftarrow self.n\_planners + 1$
\IF{$planner.score < self.best\_score$}
    \STATE $self.best\_score \leftarrow planner.score$
    \STATE $self.best\_planner \leftarrow planner$
\ENDIF
\IF{$\text{length}(self.planners) = self.reset\_period$}
    \STATE $self.generation\_no \leftarrow self.generation\_no + 1$
    \IF{$self.generation\_no < self.max\_generations$}
        \STATE $self.\text{reset\_planner\_db}()$
    \ENDIF
\ENDIF
\end{algorithmic}
\end{algorithm}

\begin{algorithm}[H]
\caption{reset\_planner\_db Method of PlannerDB Class}
\label{alg:reset_planner_db}
\begin{algorithmic}
\STATE \textbf{Input:}
\STATE \hspace{1em} $self$: PlannerDB object
\STATE \textbf{Output:}
\STATE \hspace{1em} Reset PlannerDB object with best planners
\STATE $best\_planners \leftarrow self.\text{replacement\_strategy}(self.n\_to\_keep)$
\STATE $self.planners \leftarrow []$
\STATE $self.best\_score \leftarrow \infty$
\STATE $self.best\_planner \leftarrow \text{None}$
\STATE $self.n\_planners \leftarrow 0$
\FORALL{$p \in best\_planners$}
    \STATE $self.\text{add\_planner}(p)$
\ENDFOR
\end{algorithmic}
\end{algorithm}

\begin{algorithm}[H]
\caption{replacement\_strategy Method of PlannerDB Class}
\label{alg:get_best_planners}
\begin{algorithmic}
\STATE \textbf{Input:}
\STATE \hspace{1em} $self$: PlannerDB object
\STATE \hspace{1em} $k$: Number of best planners to return
\STATE \textbf{Output:}
\STATE \hspace{1em} List of top k planners
\STATE $sorted\_planners \leftarrow \text{sort\_by\_score}(self.planners)$
\STATE \textbf{Return:} $sorted\_planners[:k]$
\end{algorithmic}
\end{algorithm}

\pagebreak
\subsection*{GenePlan}
The GenePlan class orchestrates the main evolutionary search process for generating planning algorithms. Algorithm \ref{alg:GenePlan_constructor} initializes the GenePlan object with a planner database, LLM, and optional seed planners. Algorithm \ref{alg:GenePlan_fit} runs the main evolutionary loop, iteratively generating new candidates via LLM sampling and calculating their fitness using Algorithm \ref{alg:evaluate_fitness}. Successful candidates are added to the \texttt{planner\_db} until generation limits are reached. Algorithm \ref{alg:evaluate_fitness} evaluates a planner's performance across multiple test problems by generating plans, validating them using PDDL parsing and plan validation, and computing plan length scores while handling failures appropriately.

\begin{algorithm}[H]
\caption{Constructor Method of GenePlan Class}
\label{alg:GenePlan_constructor}
\begin{algorithmic}
\STATE \textbf{Input:}
\STATE \hspace{1em} $planner\_db$: PlannerDB object
\STATE \hspace{1em} $llm$: Large language model object
\STATE \hspace{1em} $seed\_planners$: List of initial planner codes
\STATE \hspace{1em} $samples\_per\_prompt$: Number of samples per LLM call
\STATE \textbf{Output:}
\STATE \hspace{1em} Initialized GenePlan object
\STATE $self.planner\_db \leftarrow planner\_db$
\STATE $self.llm \leftarrow llm$
\STATE $self.seed\_planners \leftarrow seed\_planners$
\STATE $self.samples\_per\_prompt \leftarrow samples\_per\_prompt$
\end{algorithmic}
\end{algorithm}

\begin{algorithm}[H]
\caption{fit Method of GenePlan Class}
\label{alg:GenePlan_fit}
\begin{algorithmic}
\STATE \textbf{Input:}
\STATE \hspace{1em} $self$: GenePlan object
\STATE \hspace{1em} $test\_problems$: List of test problems
\STATE \hspace{1em} $time\_limit$: Maximum time for search
\STATE \textbf{Output:}
\STATE \hspace{1em} Fitted GenePlan object
\FORALL{$code \in self.seed\_planners$}
    \STATE $planner \leftarrow \text{Planner}(code)$
    \STATE $scores\leftarrow self.\text{evaluate\_fitness}(planner, test\_problems)$
    \STATE $planner.\text{add\_score}(scores)$
    \STATE $planner.\text{add\_feedback}()$
    \STATE $self.planner\_db.\text{add\_planner}(planner)$
\ENDFOR
\WHILE{$self.planner\_db.generation\_no < self.planner\_db.max\_generations$}
    \STATE $prompt \leftarrow self.planner\_db.\text{get\_prompt}()$
    \STATE $samples \leftarrow self.llm.\text{draw\_samples}(prompt, self.samples\_per\_prompt)$
    \FORALL{$sample \in samples$}
        \STATE $planner \leftarrow \text{Planner}(sample)$
        \STATE $scores \leftarrow self.\text{evaluate\_fitness}(planner, test\_problems)$
        \STATE $planner.\text{add\_score}(scores)$
        \STATE $planner.\text{add\_feedback}()$
        \STATE $self.planner\_db.\text{add\_planner}(planner)$
    \ENDFOR
\ENDWHILE
\end{algorithmic}
\end{algorithm}

\begin{algorithm}[H]
\caption{evaluate\_fitness Method of GenePlan Class}
\label{alg:evaluate_fitness}
\begin{algorithmic}
\STATE \textbf{Input:}
\STATE \hspace{1em} $self$: GenePlan object
\STATE \hspace{1em} $planner$: Planner object
\STATE \hspace{1em} $test\_problems$: List of test problems
\STATE \textbf{Output:}
\STATE \hspace{1em} Dictionary of scores per test
\STATE $scores\_per\_test \leftarrow \{\}$
\STATE $domain\_str \leftarrow self.planner\_db.domain$
\FORALL{$i, problem \in \text{enumerate}(test\_problems)$}
    \STATE $task \leftarrow \text{Task}(domain\_str, problem)$
    \STATE $reader \leftarrow \text{PDDLReader}()$
    \IF{$planner.\text{get\_plan}(task)$ succeeds}
        \STATE $plan \leftarrow planner.\text{get\_plan}(task)$
        \IF{$reader.\text{parse\_plan\_string}(task.problem, plan)$ succeeds}
            \STATE $parsed\_plan \leftarrow reader.\text{parse\_plan\_string}(task.problem, plan)$
            \STATE $validator \leftarrow \text{PlanValidator}()$
            \STATE $val\_result \leftarrow validator.\text{validate}(task.problem, parsed\_plan)$
            \IF{$val\_result.status = \text{VALID}$}
                \STATE $score \leftarrow \text{length}(parsed\_plan.actions)$
                \STATE $failure\_message \leftarrow \text{None}$
            \ELSE
                \STATE $score \leftarrow \text{FAILURE\_VALUE}$
                \STATE $failure\_message \leftarrow val\_result.reason.name$
            \ENDIF
        \ELSE
            \STATE $score \leftarrow \text{FAILURE\_VALUE}$
            \STATE $failure\_message \leftarrow \text{parse\_error}$
        \ENDIF
    \ELSE
        \STATE $score \leftarrow \text{FAILURE\_VALUE}$
        \STATE $failure\_message \leftarrow \text{planner\_error}$
    \ENDIF
    \STATE $scores\_per\_test[i] \leftarrow score$
    \IF{$failure\_message \neq \text{None}$}
        \STATE $planner.\text{add\_plan\_failure\_message}(failure\_message)$
    \ENDIF
\ENDFOR
\STATE \textbf{Return:} $scores\_per\_test$
\end{algorithmic}
\end{algorithm}

\end{document}